\documentclass[sigplan,screen]{acmart}
\AtBeginDocument{%
  }

\usepackage{xcolor, soul}
\usepackage{bbm}
\usepackage[table]{xcolor}
\usepackage{placeins}
\usepackage{subcaption}

\setcopyright{acmlicensed}
\copyrightyear{2026}
\acmYear{2026}
\acmDOI{XXXXXXX.XXXXXXX}
\acmConference[IEEE/ACM CHASE '26]{Conference on connected Health: Applications, Systems, and Engineering Technologies}{August 04--06,
  2026}{Pittsburgh, PA}
\begin{document}

% \title{Toward Trustworthiness of Nonparametric Deep Survival Models for Alzheimer's Disease Progression Analysis}
% \title{Are Nonparametric Deep Survival Models Trustworthy for Alzheimer's Disease Progression Analysis?}

\title[Trustworthiness of Deep Survival Models]{
Investigating Trustworthiness of Nonparametric Deep Survival Models for Alzheimer's Disease Progression Analysis}

% \affiliation{%

\author{Jacob Thrasher}
\affiliation{
\institution{West Virginia University}
\city{Morgantown}
% \state[West Virginia]
\country{USA}}
\email{jdt0025@mix.wvu.edu}

\author{Kaitlyn Heintzelman}
\affiliation{\institution{West Virginia University}
\city{Morgantown}
% \state[West Virginia]
\country{USA}}
\email{keh00023@mix.wvu.edu}

\author{Peter Martone}
\affiliation{\institution{West Virginia University}
\city{Morgantown}
% \state[West Virginia]
\country{USA}}
\email{pmartone@mix.wvu.edu}

\author{David Kotlowski}
\affiliation{\institution{West Virginia University}
\city{Morgantown}
% \state[West Virginia]
\country{USA}}
\email{dtk00010@mix.wvu.edu}

\author{Binod Bhattarai}
\affiliation{\institution{University of Aberdeen}
\city{Aberdeen}
% \state[West Virginia]
\country{Scotland}}
\email{binod.bhattarai@abdn.ac.uk}

\author{Donald Adjeroh}
\affiliation{\institution{West Virginia University}
\city{Morgantown}
% \state[West Virginia]
\country{USA}}
\email{donald.adjeroh@mail.wvu.edu}

\author{Prashnna Gyawali}
\affiliation{\institution{West Virginia University}
\city{Morgantown}
% \state[West Virginia]
\country{USA}}
\email{prashnna.gyawali@mail.wvu.edu}

\renewcommand{\shortauthors}{Thrasher et al.}

\begin{abstract}
Alzheimer's Dementia (AD) is a progressive neurodegenerative disease
marked by irreversible decline, making reliable modeling of its progression essential for effective patient care.
Progression-aware methods such as survival analysis are therefore crucial tools for the early detection and monitoring of AD. Recent advancements in deep learning have demonstrated remarkable performance in survival tasks, but alarmingly fewer studies have been conducted in the domain of AD. Further, the studies that do exist do not consider learned bias within the model itself, which could result in unfair and unreliable predictions toward certain marginalized groups. As such, we conduct a rigorous study of fairness in AD progression analysis along with a thorough feature importance study to determine the characteristics which are most important for reliable AD predictions. Furthermore, we propose two novel fairness metrics, called Time-Dependent Concordance Impurity and Kaplan-Meier Fairness, to quantify bias with respect to sensitive attributes such as sex, race, and education in nonparametric survival models. Our study demonstrates that while deep learning powered survival models are robust tools which can aid clinicians in AD care decisions, they often exhibit considerable bias, representing important avenues for future research.
\end{abstract}







\maketitle

\section{Introduction}

Alzheimer’s disease (AD) is a progressive neurodegenerative disorder and the most common cause of dementia in older adults, affecting nearly one in three individuals aged 85 years and older \cite{AlzheimersAssociation2024FactsFigures}. AD is characterized by irreversible cognitive decline, including short-term memory loss and a gradual loss in independence. A key challenge in Alzheimer’s care is the substantial heterogeneity in disease progression rates across individuals, even among patients with similar clinical profiles \cite{Duara2022-bt}. This unpredictability complicates clinical decision-making, long-term care planning, and patient counseling. Additionally, current clinical assessments often identify AD only after substantial neurological damage has occurred, limiting eligibility for disease-modifying therapies and reducing the potential impact of intervention \cite{Sperling2013-bt, Nakashima2025-yy}. 

These challenges motivate the need for predictive frameworks that models not only whether AD will occur, but when clinically meaningful transitions are likely to take place. This challenge naturally aligns with the framework of survival analysis, which aims to predict the time until the occurrence of a specific event and is particularly well suited to degenerative conditions like AD, where disease processes are irreversible and occur over an extended time frame.
%such as death, relapse, or onset dementia  \cite{klein2013survival, clark2003survival, AD-time-to-event, Abuhantash2025, Huh2020}. 
%This approach is particularly well suited to degenerative conditions like AD, where disease processes are irreversible and occur over an extended time frame.  
In contrast to traditional classifiers that detect AD only after progression has occurred, survival models enable estimation of an individual's risk trajectory, such as the time to conversion from a non-AD state to AD diagnosis. By modeling disease progression over time, survival analysis supports  more informed clinical decision-making and enables proactive intervention aimed at slowing disease progression.

While traditional survival modeling techniques have seen success, modern advancements in deep learning (DL) have resulted in rapid growth in high-dimensional tasks \cite{wiegrebe2024deep}.
% proved to be a desirable alternative due to their ability to capture more complex features from high-dimensional signals \cite{wiegrebe2024deep}. 
These advancements have enabled researchers to model disease progression in AD \cite{thrasher2024tessltimeeventawareself, AD-time-to-event, Abuhantash2025, Huh2020}, cancer \cite{saeed2024survrnclearningorderedrepresentations, xu2025distilledpromptlearningincomplete}, embolism \cite{huo2025timetoeventpretraining3dmedical}, and relapse \cite{clark2003survival} tasks, highlighting its effectiveness in critical domains. 
Moreover, DL has facilitated the development of nonparametric deep survival models (NDSMs), which can effectively produce individualized survival distributions (ISD) without imposing restrictive modeling assumptions (detailed further in Sec \ref{sec:related}). While NDSMs have demonstrated strong discriminative performance in survival modeling  \cite{thrasher2024tessltimeeventawareself, Tang_Zhang_Li_2025}, 
existing studies largely confine their evaluation to performancec-centric metrics.
In particular, most prior work emphasizes discrimination—often measured solely through the Concordance Index \cite{Antolini2005ATD}—while overlooking other critical aspects of a robust survival modeling framework, such as calibration, fairness, and interpretability.
Meanwhile, other lines of work have begun to explicitly investigate properties such as interpretability \cite{survshap, KRZYZINSKI2023110234} and fairness \cite{zhang2022longitudinalfairnesscensorship} in survival modeling. However, these studies are typically validated on synthetic data or relatively low-complexity benchmark datasets, rather than on heterogeneous, high-dimensional real-world clinical cohorts. As a result, it remains unclear whether their proposed approaches retain these desirable properties when applied to challenging real-world settings such as AD progression modeling.

% these works often neglect other important aspects of a holistic evaluation of the survival framework, including calibration, fairness, and interpretability. 

% Discrimination alone is insufficient to guarantee the efficacy of a proposed method as it inadvertently conceals flaws in other areas. This severely limits the trustworthiness of NDSMs and therefore their translational ability. 

% To this end, we believe that it is currently inconclusive whether NDSMs are suitable for complex real-world tasks such as AD progression analysis. Existing work in AD progression often limits analysis
% to measures of discriminative performance through the Concordance Index \cite{Antolini2005ATD} alone. 

% Meanwhile other works which aim to explicitly evaluate model interpretability \cite{survshap, KRZYZINSKI2023110234} and fairness \cite{zhang2022longitudinalfairnesscensorship} do not typically analyze such metrics on highly complex real world data. Rather, they primarily demonstrate their theoretical success on synthetic or low-complexity benchmark datasets.

Motivated by this, we propose a comprehensive evaluation pipeline for trustworthiness in AD progression modeling. This pipeline examines NDSM suitability in real world tasks by providing a well-rounded framework for analyzing model performance. Additionally, we claim that established fairness metrics \cite{zhang2022longitudinalfairnesscensorship, zhang-fairness} are not suitable for nonparametric survival models, and introduce the \textit{Time-Dependent Concordance Impurity (CI-td)} and \textit{KM-Fair Calibration} metrics to quantify the bias of NDSMs. Formally, we present our contributions as follows: (1) We introduce two novel bias quantification metrics to allow for the evaluation of fairness in nonparametric deep survival models. (2) We present a rigorous evaluation pipeline to analyze NDSM suitability for real-world AD progression modeling by analyzing discrimination, calibration, fairness, and interpretability. (3) We conduct an extensive analysis of bias in deep survival models across sex, ethnicity, and education to evaluate the fairness of state of the art models.

\section{Related Work}
\label{sec:related}
\subsection{Survival analysis}
Parametric methods assume that the survival function adheres to a specific prescribed distribution such as Weibull \cite{deepweisurv}, Log Normal \cite{lognorm-mle}, or Asymmetric Laplace \cite{sheng2025learning}. These models then aim to estimate the the associated parameters and sample ISDs from the learned distributions.

Meanwhile, semi-parametric methods such as DeepSurv \cite{Katzman_2018} instead operate under the \textit{proportional hazards} assumption, 
% Deep survival analysis was first proposed by Katzman et al in 2018 with their DeepSurv \cite{Katzman_2018} algorithm. DeepSurv is a \textit{proportional hazard} model, 
which assumes that there exists some "baseline hazard function" $h_0$, which describes the average hazard of the population. Proportional hazards models then predict a scalar value, known as the \textit{hazard coefficient}, that linearly scales $h_0$ to create the ISD for a given patient.

While these methods saw some success, more recent work instead aims to directly predict individualized survival distributions in a nonparametric manner. This is typically achieved by decomposing the time horizon into $T$ discrete bins and estimating the probability of event occurrence at each interval. Under this formulation, Negative Log-Liklihood (NLL) can be extended to support right-censored data \cite{lee2024toward, deephit}. Similarly, Ranked Probability Scoring (RPS) \cite{rps} also predicts the event probability at each time interval but instead aims to improve model calibration by simultaneously considering the entire time horizon. While these methods sacrifice some fine-grained survival information by binning the study window, they achieve superior performance due to their ability to model survival distributions at the individual level.

% For example, DeepHit \cite{deephit} decomposes the time horizon into $T$ discrete intervals and estimates the probability of event occurrence at each window via negative log-likelihood. Similarly, Ranked Probability Scoring (RPS) \cite{rps} also predicts the event probability at each time interval but instead aims to improve model calibration by simultaneously considering the entire time horizon. While these methods sacrifice some fine-grained survival information by binning the study window, they achieve superior performance due to their ability to model survival distributions at the individual level.

Beyond theoretical contributions to deep learning, the medical community has worked to apply these survival models to Alzheimer's Dementia disease progression. Huh et al. \cite{Huh2020} apply traditional (non-DL) methods to data from the Korean National Health Insurance Service to quantify the risk of patients from different population subgroups. Meanwhile, \cite{Abuhantash2025} conducts a comprehensive feature importance study from Alzheimer's data provided by the Alzheimer's Disease Neuroimaging Initiative (ADNI). Finally, \cite{AD-time-to-event} evaluates the performance of deep survival models on data from the National Alzheimer's Coordinating Center (NACC) and apply recursive feature elimination to determine the most important features for AD progression prediction. Notably, none of these works utilize NDSM framework for their analysis.

\subsection{Fairness}

\textit{Fairness} refers to the process of ensuring that a deep model evaluates all subgroups of a population equally. These subgroups can be stratified based on sensitive attributes such as sex or ethnicity. A "fair" model should then behave similarly across each group, attaining similar evaluation metrics \cite{zhang2022longitudinalfairnesscensorship}. 
\textit{Bias} refers to the task of quantifying the model's bias toward any one specific group. This concern is particularly relevant in AD, where disparities in diagnosis, progression rates, and access to care across demographic groups have been well documented, raising the risk that biased models may further amplify existing inequities \cite{Mayeda2016-qb, Lewis2023-xb}.

There exists rich literature which studies the topic of bias and fairness in machine learning \cite{Goktas2025, Yang2024DemographicBOA, Liu2023ATPA}, but these methods typically rely on the presence of a class label -- a feature which is notably absent in the survival framework due to data censorship. As such, little work address the issue of censorship as it pertains to bias and fairness. Further, the established metrics \cite{zhang2022longitudinalfairnesscensorship, zhang-fairness} are not suitable for nonparametric survival models due to their reliance on hazard ratios generated by semi-parametric methods, representing an important gap in the literature.

% Bias quantification is particularly challenging in survival analysis due to the presence of censored data. To address this, \cite{zhang2022longitudinalfairnesscensorship} propose metrics to evaluate \textit{group fairness} (model fairness across group categories such as sex and ethnicity). One such metric, Concordance Impurity (CI) measure the model's capability to properly rank the hazard of comparable pairs in a subgroup. An unbiased deep survival model will rank pairs across different population groups with relatively similar accuracy, whereas a biased one might favor one group over another. This work is extended in \cite{zhang-fairness} to evaluate \textit{individual fairness} whereby the model must treat individuals who are similarly situated (i.e. have similar features) similarly in inference. This is evaluated through Discounted Cumulative Fairness, which ensures that the representation space aligns with proper ranking order of comparable individuals.
\section{Method}

\subsection{Problem formulation}

We consider the task of Alzheimer's Dementia progression modeling as a time-to-event prediction problem, where we are interested in estimating the likelihood of a non-AD subject to receive a clinical AD diagnosis at a given point in time. Consider a longitudinal study which monitors $N$ potential AD patients over some time horizon. We construct a survival dataset $\mathcal{D} = \{X_i, \delta_i, t_i\}_{i=0}^N$, where $X_i$ is a clinical feature vector, $\delta_i$ is the event indicator, and $t_i$ is the time until event observation or censorship for some subject $i$. Specifically, $\delta=1$ if the event was observed within the study window and $\delta=0$ otherwise. The latter is referred to as \textit{right-censorship}, and is an integral component of the survival modeling process because it is unknown if/when the event will occur in the future. In these cases, we record the last visit as the time of censorship. 

The goal is then to learn a function $f_\theta(X_i)$ which estimates an individualized survival distribution $S_\theta(t|X_i) = P(T>t)$, representing the probability of a subject \textit{not} receiving a clinical AD diagnosis by time $t$. The primary benefit of utilizing NDSMs is their ability to directly learn $S_\theta(t|X_i)$ from the data without imposing strong distributional assumptions or using proportional hazards. However, this process also reduces the overall trustworthiness of the survival model as it is more difficult to truly understand decision making dynamics then their (semi)parametric counterparts. For this reason, NDSMs require a more robust evaluation framework to validate their efficacy in real world scenarios. We argue that a well rounded analysis of an NDSM should evaluate its \textit{discrimination, calibration, fairness} and \textit{interpetability}. In this section, we outline each of these categories.

\subsection{Quantifying NDSM performance}
Discrimination measures the model's ability to properly rank elements in order of risk. If subject A is known to observe an event before subject B, then a properly trained survival model should rank their risks accordingly. However, this does not provide guarantees that the model's predictions actually align with the true underlying survival distributions.
Instead, this quality is measured through model \textit{calibration} \cite{lee2024toward, pmlr-v235-qi24a, qi2024toward}. Importantly, strong performance in discrimination does not imply well aligned calibration, and vice versa. In fact, there is often a trade-off between these two objectives, known as the \textit{discrimination-calibration tradeoff}, where one metric weakens while the other improves \cite{pmlr-v235-qi24a}. For this reason, it is imperative that NDSMs are evaluated through a combination of discrimination and calibration metrics.

\subsubsection{Discrimination}
The Time-dependent Concordance Index (C-td) \cite{Antolini2005ATD} measures the ratio of correctly ordered elements to the number of total comparable pairs. A pair of elements $i, j$ are considered comparable if element $i$ is known to be observed before $j$, or formally, $\delta_i=1$ and $t_i < t_j$. 

We can then express the model's belief that an event will occur \textit{at or before} some time $t$ via the Cumulative Incidence Function (CIF), defined as:

% To compute the C-td, we first must define the Cumulative Incidence Function (CIF), which expresses the probability that the event will occur at or before some time $t$. This is essentially the cumulative sum of the $S(t|X)$, and is defined as:

\begin{equation}
    F(t|X) = \sum^{t}_{i=0} \text{softmax}(f_\theta(X))_i
\end{equation}

where $f_\theta(X)$ is a deep neural network parameterized by $\theta$, and $t$ is the time horizon. Finally, C-td is computed as:

\begin{equation}
    \text{\textit{C-td}} = \frac{A_{i, j} * \mathbbm{1}(F(t_i|X_i) > F(t_i|X_j))}{\sum_{i \neq j} A_{i, j}}
\end{equation}

Here, $A_{i, j}$ is the set of all comparable elements, defined as $A_{i, j} = \mathbbm{1}(\delta_i = 1, t_i < t_j)$. Higher is better for C-td, where 1 corresponds to perfect ranking, 0.5 is complete randomness, and 0 is inversely perfect.

\subsubsection{Calibration}
The Kaplan-Meier (KM) curve \cite{Kaplan01061958} represents the general population-level survival distribution of a group and can easily be calculated non-parametrically with only time and event labels. As such, it serves as a good target distribution for model alignment since it is impossible to obtain ground truth ISDs. KM-Cal ($K$) \cite{XCAL} measures calibration by computing the KL-divergence between the Kaplan-Meier curve and the average of all ISD predictions from the test set, defined as:

\begin{equation}
    K = KL(S_{KM}(t) || \hat{S}(t))
\end{equation}

%However, for our implementation, we chose to adapt a modified version which instead measures the average Mean Squared Error for numerical stability \cite{pmlr-v235-qi24a}:

% \begin{equation}
%     K = \frac{1}{t_{max}}\sum_{t=0}^{t_{max}} (S_{KM}(t) - \hat{S}(t))^2dt
% \end{equation}

where $S_{KM}$ is the survival distribution predicted by the KM estimator and $\hat{S}(t)$ is the average of all survival distributions predicted by $f_\theta(\cdot)$. If these distributions are similar (small KL divergence value), then the model is likely to be calibrated. 

\subsubsection{Integrated Brier Score}
The Brier Score (BS) is a metric which evaluates the model's accuracy at a specific point in time $t$. This is effectively quantified as the mean squared error between the survival prediction $\hat{S}(t|X)$ and 1 if the event was observed at $t$ and 0 otherwise. This formulation can be further decomposed into three additive components: $BS = CAL - RES + UNC$ \cite{ANewVectorPartitionoftheProbabilityScore}. Here, \textit{CAL} refers to the model's calibration. \textit{RES} (resolution) measures how much the conditional probabilities deviate from the prediction average, essentially penalizing the model for repeating the same probabilities across many samples. Finally, \textit{UNC} refers to the overall uncertainty of the prediction. Thus, the Brier Score cannot explicitly be considered a measure of discriminative performance, nor calibration. Instead it represents an aggregation of overall performance.

The Integrated Brier Score (IBS) simply calculates the average BS across all intervals in the time horizon. IBS can take on any value in a range [0, 1], where lower is better. \cite{pysurvival_cite}.

\subsection{Evaluating fairness}

\subsubsection{Time-Dependent Concordance Impurity}

% \begin{figure}[t]
% \label{fig:CI}
%   \centering
%    \includegraphics[width=\linewidth]{figures/CI.png}

%    \caption{Example of Concordance Impurity calculation.}
%    \label{fig:CI}
% \end{figure}

Concordance Impurity (CI) \cite{zhang2022longitudinalfairnesscensorship} measures the variation in discriminative performance for different subgroups of a population. For example, consider the sensitive attribute "sex".
% Figure \ref{fig:CI} provides an example with the sensitive attribute sex. 
We first partition the dataset into male and female populations, then calculate the overall concordance with respect to each group. This value is called the \textit{concordance fraction (CF)}. However, the original authors \cite{zhang2022longitudinalfairnesscensorship} utilize risk scores to determine the concordance of two elements. This limits Concordance Impurity to parametric and semi-parametric methods as NDSMs do not have a notion of "risk". To adapt CI to the nonparametric framework, we instead evaluate the concordance directly through the predicted survival distribution. Specifically, we consider an $(i, j)$ patient pair to be concordant if $\hat{S}(t_i|X_i) < \hat{S}(t_i|X_j)$, so long as such pair is comparable. This allows us to extend the impurity metric to function with nonparametric survival models, allowing for the fairness evaluation of NDSMs. Once the CF is calculated for each group, the largest difference in CF across all subgroups is returned as the impurity score:

\begin{equation}
    \text{CI-td} = \text{min}\{CF_{g_i} - CF_{g_j}|i\neq j\}
\end{equation}

where $g_i, g_j$ refer to the groups specified by the sensitive attribute. 

\subsubsection{KM Fair Calibration}
Similar to Concordance Impurity, calibration fairness can be calculated by assessing variation in calibration across subgroups of a population. Previous work \cite{zhang2022longitudinalfairnesscensorship} have utilized the Hosmer-Lemeshow goodness-of-fit statistic \cite{Hosmer01011980} to analyze the agreement between predictions and observed outcomes. This results in a clean decision rule which can be used to classify models as "fair-" or "biased-calibrated", defined as:

\begin{equation}
    HL_g(S(t|X)) = \sum_{i=0}^{t_{max}} \frac{(KM_{ig} - p_{ig}^2)n_{ig}}{p_{ig}(1-p_{ig})}
\end{equation}

where $g$ is the group identifier, $p_i$ is the predicted probability at time $i$ and $n_i$ is the number of observations at $i$. However, this statistic is scaled by the number of event observations at a given time interval. For highly underrepresented populations which may only have a limited number of elements in the dataset, this could lead to greater statistical uncertainty \cite{zhang2022longitudinalfairnesscensorship}. Instead, we chose to develop a novel metric, \textbf{KM-Fair}, which extends KM-Cal for fairness evaluation.

To do this, we compute the group-wise KM distribution $KM_g(t)$ and corresponding model predictions $\hat{S}(t)$ to determine the KM-Cal score for each group in the population. To improve the statistical certainty of the individual scores, we bootstrap the process by repeating the calculation $B$ times with random subsets of the group population. This results in a vector $\vec{K}_{g_i}$ of $B$ KM-Cal scores for each group $g_i$. We use a value of $B=1000$ in our experiments.

% This results in a set $G=\{K_{g_1}, K_{g_2}, ..., K_{g_m}\}$ of calibration scores for $m$ groups. 

We perform pair-wise bias analysis by measuring the difference between the bootstrapped group KM-Cal scores between two populations. Specifically, let $\vec{K}_{diff} = \vec{K}_{g_i} - \vec{K}_{g_j}$ be a vector which contains the difference in bootstrapped calibration scores between two groups. If the model exhibits more bias toward $g_i$, then $\vec{K}_{diff}$ be almost entirely negative. Similarly, if there is substantially more bias toward $g_j$, then $\vec{K}_{diff}$ will be mostly positive. With this in mind, we compute the 95\% confidence interval $[a, b]$ for $\vec{K}_{diff}$ and formalize the decision rule as follows:

\begin{equation}
\label{eq:km-fair}
\text{KM-Fair}=
\begin{cases}
    -1, & a, b < 0 \\
    0,  & a \leq 0 \leq b \\
    1, & 0 < a, b
\end{cases}
\end{equation}

This formulation provides insights into the model's overall fairness as well as to which direction it tends to exhibit bias.

\subsection{Interpreting NDSM predictions}
To improve model interpretability and assess the robustness of the learned representations, we performed permutation-based feature importance analysis on our trained survival models. After model training, the best-performing checkpoint for each model, selected according to the validation concordance index, was loaded and evaluated on a held-out test set. Baseline predictive performance was measured using the C-td. Feature importance was estimated by independently permuting each input feature across samples within the test set while leaving all other features unchanged. This permutation preserves the marginal distribution of feature values but disrupts their correspondence with individual samples. The trained model was then re-evaluated on the permuted test set, and the change in C-td relative to baseline performance was recorded. To reduce variability introduced by random permutations, the procedure was repeated 10 times per feature using independently generated permutations. The mean decrease in C-td across repetitions was used as the feature importance score, where larger decreases indicate greater dependence of model performance on the corresponding feature. All evaluations were conducted under identical inference settings to ensure that performance changes were attributable solely to feature permutation.

% \subsection{Survival models}
% In this study, we evaluate the comprehensive performance of the most popular NDSM frameworks in recent literature. Specifically, we analyze Negative Log-Likelihood (\textbf{NLL}) \cite{pycox}, Neural Multi-Task Logistic Regression (\textbf{N-MTLR}) \cite{MTLR}, and Ranked Probability Scoring (\textbf{RPS}) \cite{rps11} techniques. Additionally, some works have explored the use of an additional ranking component, which seeks to improve performance by learning the proper ordering of uncensored individuals. As such, we include \textbf{DeepHit} ($\mathcal{L}_{NLL} + \mathcal{L}_{Ranking}$) \cite{deephit} and \textbf{RPS+Rank} ($\mathcal{L}_{RPS} + \mathcal{L}_{Ranking}$) \cite{rps11} in our suite yielding a total of \textbf{5} methods for evaluation.

\section{Data}

\subsection{Survival data}

\begin{figure*}[t]
  \centering
   \includegraphics[width=.8\linewidth]{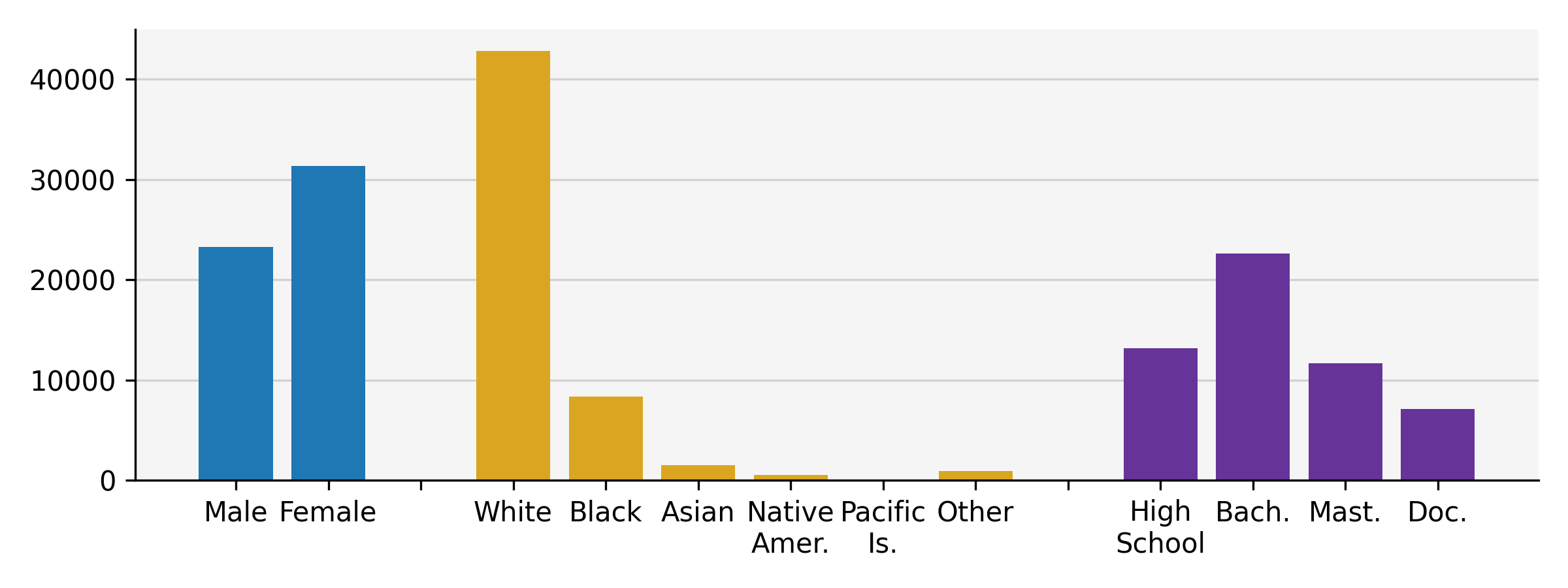}

   \caption{Breakdown of NACC demographic characteristics use for NDSM fairness evaluation. }
   \label{fig:demo}
\end{figure*}

We utilize follow-up data from the National Alzheimer's Coordinating Center (NACC) \cite{Kukull2025}. This dataset contains over 200,000 records from 55,000 unique patients. Within this, there are approximately 1,000 unique features which encompass demographic, clinical, cognitive, and diagnosis information. Of these features, we identify \textit{three} sensitive attributes which we use for our fairness analysis: \textit{sex, race}, and \textit{education}. Figure \ref{fig:demo} provides a breakdown of the population representation for each of these attributes. 

Notably, NACC does not explicitly include the proper channels to model survival information, namely event indicator and time labels. 
We first define the the positive event to be the case in which a given patient \textit{converts} from a non-AD diagnosis to AD. Then, the survival time is recorded as the difference between their initial visit and the visit in which they received an AD diagnosis.
For example, a patient with Mild Cognitive Impairment (MCI) at month 0 that later converts to AD at month 12, will be recorded as the tuple $(X, \delta=1, t=12)$. Otherwise, if the patient still has not received an AD diagnosis by their last visit, we consider them to be censored and record their time of censorship in a similar manner. Since NDSM model survival over discrete intervals across the time horizon, we then binned the continuous time labels into 10 intervals using the pipeline laid out in \cite{pycox}.

One such challenge with defining the survival data for the NACC dataset is that the only definitive way to determine an AD diagnosis is post-mortem. This means that patient diagnosis is not always definitive and may oscillate between positive and negative diagnoses. To account for this, we take the \textit{last known} diagnosis as the final, most accurate diagnosis label for determining converters. With this approach, a subject who receives a positive AD diagnosis at one month but later is changed to non-AD at a later time will be recorded as censored for the purpose of our data.

\subsection{Feature selection}

While deep learning facilitates high-dimensional survival modeling, it is often convention to exclude irrelevant and redundant features from the feature set to create a smaller, more manageable environment for the survival model. As such, we manually selected a subset of 98 features which capture demographic, genetic, behavior, risk, and clinical assessment data to construct our survival dataset. Further details can be found in Appendix \ref{app:feat}. These features were selected based on relevance to Alzheimer's Dementia diagnosis and alignment to other AD studies \cite{Sharma2021-ll}. 

\subsection{Data processing}
Our data processing pipeline consists of 3 main steps (1) truncation of survival data (2) data imputation, and (3) normalization and one-hot encoding

\underline{Truncation:} Generally speaking, survival data should only contain patients who are ``at risk" during observation. This means that subjects with an observed event \textit{at their baseline visit} should not be considered for the purposes of the study. This is because these elements could introduce bias in the survival modeling process. To understand this, consider two subjects who received an AD diagnosis at their baseline visit (i.e. $\delta_i=1, t_i=0$). These elements can be thought of as being \textit{left-censored}, meaning that their event time occurred before the study window. We will consider these cases as having ``negative" time. Therefore, if subject A developed AD 1 month prior to their baseline visit ($t=-1$) and subject B 12 months prior ($t=-12$), the survival model will still treat both subjects as at the same risk, despite subject B being further in progression than subject A. For this reason, we truncate the survival data by dropping all elements where $\delta_i=1, t_i=0$.

\underline{Imputation:} There were missing elements from many of the selected features in our dataset. A common method to handle such cases is through \textit{data imputation}, where missing elements are replaced with estimates (usually the mean or mode of the feature) to ensure stability during training. As such, we imputed all continuous valued features with the mean value of all available elements and used the mode for categorical features. To avoid excessive imputation, we also dropped all features that we missing more than 30\% of their values.

\underline{Normalization and one-hot encoding:} We finalize our preprocessing pipeline by normalizing all continuous value features to be in range [0, 1] through the min-max normalization scheme:

\begin{equation}
    X^* = \frac{X - X_{min}}{X_{max} - X_{min}}
\end{equation}

We then one-hot encoded non-binary categorical variables (i.e. race) into binary categories.

% \subsection{Implementation}
% Beyond data preparation, we have implemented a full deep survival analysis pipeline. While this pipeline is currently only compatible with benchmark survival datasets that are readily available through API calls, we are working to support external datasets such as NACC. Simultaneously, we are working to implement the Concordance Impurity (CI) as discussed in Sec. \ref{sec:related}. Unfortunately, the authors did not release the code with their article, so the metric must be implemented from scratch. Once completed, we will be able to completely train and evaluate model bias on our Alzheimer's data.
\section{Experimental Setup}

We evaluate the survival and fairness performance across a suite of commonly used state-of-the-art NDSMs.
These models include Negative Log-Liklihood (\textbf{NLL}) \cite{pycox}, Neural Multi-Task Logistic Regression (\textbf{N-MTLR}) \cite{MTLR}, \textbf{DeepHit} \cite{deephit} and Ranked Probability Scoring (\textbf{RPS}) \cite{rps11}, and \textbf{RPS+Rank} \cite{rps11}. DeepHit and RPS+Rank utilize an additional ranking loss component which adapts the idea of Concordance to learn the proper ordering of uncensored individuals and can be expressed as $\mathcal{L}_{DeepHit} = \mathcal{L}_{NLL} + \mathcal{L}_{Ranking}$ and $\mathcal{L}_{RPS+Rank} = \mathcal{L}_{RPS} + \mathcal{L}_{Ranking}$, respectively.

All models were trained using the Adam optimizer \cite{kingma2017adammethodstochasticoptimization} with a learning rate of $1 \times 10^-4$ and a batch size of 128 for 20 epochs. We present our results as an average across three random seeds, with the corresponding standard deviations included in parenthesis. 

% For a comprehensive analysis, we evaluate the survival performance and fairness of \textbf{3} state of the art survival models, including Negative Log-Likelihood (\textbf{NLL}) \cite{pycox}, Neural Multi-Task Logistic Regression (\textbf{N-MTLR}) \cite{MTLR}, Ranked Probability Scoring (\textbf{RPS}) \cite{rps11}. Additionally, some works have explored the use of an additional ranking component, which seeks to learn the proper ordering of uncensored individuals. As such, we include \textbf{DeepHit} ($\mathcal{L}_{NLL} + \mathcal{L}_{Ranking}$) \cite{deephit} and \textbf{RPS+Rank} ($\mathcal{L}_{RPS} + \mathcal{L}_{Ranking}$) \cite{rps11} in our suite yielding a total of \textbf{5} methods for evaluation.

\section{Results}

% \begin{table*}[]
% \centering
% \small
% \begin{tabular}{@{}lcccc@{}}
% \toprule
% \textbf{Model} & \textbf{C}              & \textbf{IBS}            & \textbf{CI-td (Sex)}    & \textbf{CI-td (Race)}   \\ \midrule
% NLL            & \textbf{0.8704 (0.001)} & \textbf{0.0864 (0.001)} & 0.0136 (0.003)          & \textbf{0.0665 (0.012)} \\
% DeepHit        & 0.8663 (0.003)          & 0.0967 (0.001)          & 0.0120 (0.008)          & 0.0994 (0.020)          \\
% MTLR           & 0.8694 (0.001)          & 0.0875 (0.001)          & 0.0228 (0.005)          & 0.0980 (0.005)          \\
% RPS            & 0.8662 (0.001)          & 0.0886 (0.001)          & 0.0066 (0.003)          & 0.0711 (0.029)          \\
% RPS + Ranking  & 0.8666 (0.001)          & 0.0887 (0.001)          & \textbf{0.0063 (0.003)} & 0.0851 (0.029)          \\ \bottomrule
% \end{tabular}
% \caption{Average survival and bias results, where the value in parenthesis indicates the standard deviation across three seeds.}
% \label{tab:results}
% \end{table*}

% \usepackage{booktabs}
\begin{table*}[]
\begin{tabular}{lcccccc}
\hline
Model        & C-td $\uparrow$                   & IBS $\downarrow$                 & KM-cal $\downarrow$              & CI-td (sex)  $\downarrow$        & CI-td (race) $\downarrow$         & CI-td (educ) $\downarrow$        \\ \hline
NLL          & \textbf{86.52 (0.06)} & \textbf{8.89 (0.06)} & .6035 (.0015)          & \textbf{2.14 (0.24)} & 17.27 (2.58)          & 6.71 (1.20)          \\
DeepHit      & 86.23 (0.04)          & 9.70 (0.16)          & .6011 (.0021) & 3.42 (0.69)          & 18.89 (4.68)          & 7.64 (1.23)          \\
N-MTLR         & 86.39 (0.02)          & 8.94 (0.09)          & .5743 (.0066)          & 2.98 (1.44)          & 18.60 (4.85)          & \textbf{4.48 (1.92)} \\
RPS          & 86.04 (0.21)          & 9.24 (0.06)          & .5598 (.0023)          & 5.42 (0.98)          & \textbf{14.88 (3.32)} & 10.67 (0.61)         \\
RPS+Rank & 86.06 (0.23)          & 9.26 (0.08)          & \textbf{.5588 (.0026)}          & 5.67 (1.18)          & 16.21 (3.20)          & 11.08 (0.86)         \\ \hline
\end{tabular}
\caption{Metrics for primary experiments. $\uparrow$ indicates higher values preferred, $\downarrow$ indicates lower values preferred. Values in parentheses represent standard deviations.}
\label{tab:main-results}

\end{table*}

\begin{figure*}[t]
\label{fig:importance}

    \centering
    \includegraphics[width=1\linewidth]{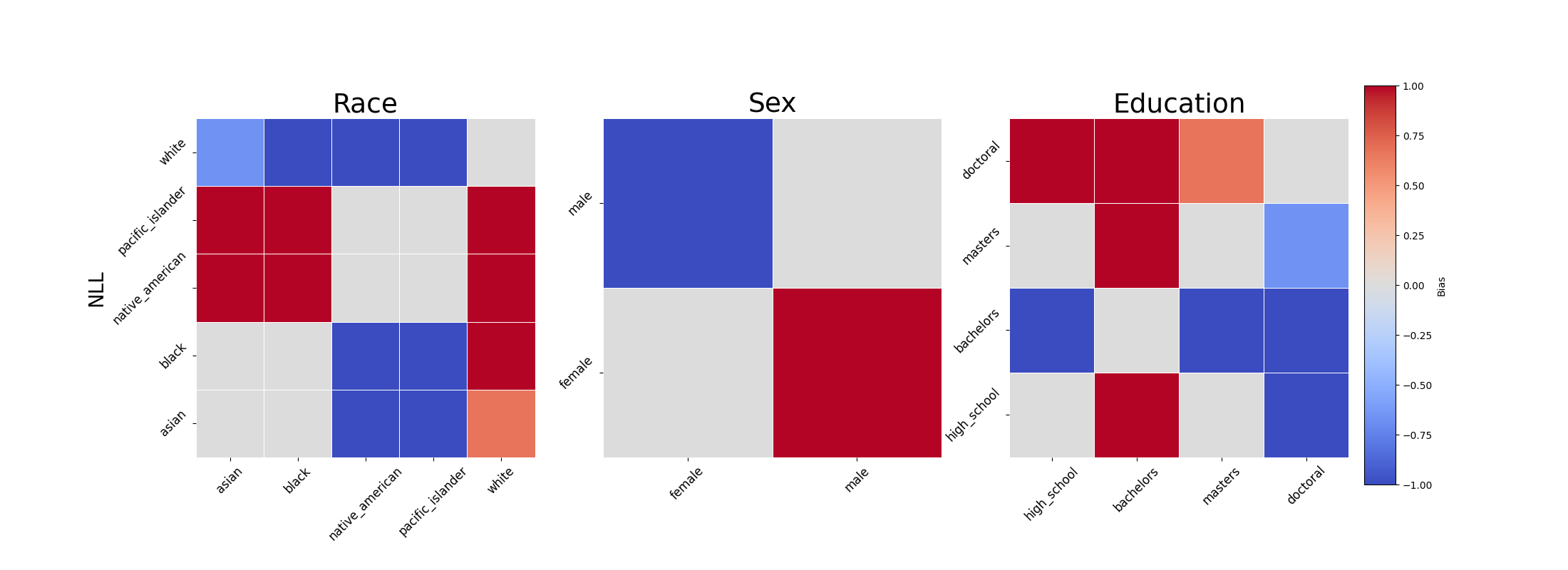}
    \caption{KM-Fair analysis for NLL model where \colorbox{blue!50}{blue} indicates a model which is biased toward \textit{row attributes} and \colorbox{red}{red} indicates one which is biased toward \textit{column attributes}.}
    \label{fig:km-fair}
\end{figure*}

We provide our primary results for survival and impurity, averaged across three random seeds, in Table \ref{tab:main-results}. Figure \ref{fig:km-fair} then shows a visualization of the pair-wise KM-Fair scores for each sensitive attribute. These matrices showcase the average decision score from Eq. \ref{eq:km-fair} across 3 seeds, where negative values (colored blue) indicate that the model tends to bias toward the \textit{row} attributes whereas positive values (colored red) correspond with a bias toward \textit{column} attributes in terms of fairness. We also note that we only include NLL as a representative example for brevity. The remaining KM-Fair plots can be found in Appendix \ref{app:KM-fair}

While all models overall performed well for the survival task, we observe that the \textit{discrimination-calibration tradeoff} is apparent. RPS, which is explicitly trained to optimize calibration, outperforms NLL, DeepHit, and N-MTLR methods in terms of calibration, but is considerably weaker in discriminative performance. Even further, across all tested models, RPS suffers the greatest instability in discrimination, being the only method with a standard deviation >0.1. Surprisingly, we also note that NLL, which is the simplest of the analyzed methods, yields the strongest discriminative survival performance. This could indicate that additional complexity introduced by the subsequent methods may not be necessary for the AD progression task.

In terms of fairness, our experiments show that all models exhibit low impurity scores for the \textit{sex} attribute. This could be due to the fact that the dataset is somewhat balanced in terms of sex ($\sim$42/48\% male/female split). Even so, from Figure \ref{fig:km-fair}, it can be seen that despite having the lowest impurity score, NLL exhibits a bias toward male subjects. 

When considering attributes with larger skew in representation such as \textit{race} and \textit{education}, NDSMs exhibit considerably more bias. Here, RPS based methods have superior impurity with respect to \textit{race}, but are simultaneously  worse in \textit{education} fairness. Meanwhile, the reverse is true for the remaining NDSMs, which demonstrate high \textit{race} impurity and low \textit{education} impurity. Additionally, the results in the KM-Fair analysis align with general intuition of bias in ML methods. It can be seen that all models are generally biased toward populations with larger representation in the dataset such as white and black patients as well as ones with Bachelors degrees.

% We also observe that all models obtain very low impurity scores for the Sex attribute, indicating that nonparametric survival models are often fair when evaluated across male and female populations. This is likely because the dataset was generally balanced between sex, which helped to prevent the model from overfitting toward one over the other. Conversely, it can be seen that all models incur some amount of bias when it comes to the Race category. This is unsurprising given that race is often much more skewed across the five categories. The dataset was predominately white, with comparatively less representation for minority groups such as Black, Asian, Native American, and Pacific Islander populations.

\subsection{Impact of sensitive attributes during training}

\begin{table*}[]
\centering
\begin{tabular}{lcccccc}
\hline
\multicolumn{7}{c}{-Sex}                                                                       \\ \hline
Model                & C-td              & IBS            & KM-cal     & CI-td (sex)    & CI-td (race)   & CI-td (educ)  \\ \hline
NLL                  & \cellcolor{red!25}-0.07          & \cellcolor{green!25} -0.09 & - .0007          & \cellcolor{green!25} -0.39 & -2.07          & 0.32          \\
DeepHit              & 0.03           & \cellcolor{green!25} -0.26          & \cellcolor{green!25} -.0022 & \cellcolor{green!25}-1.06          & -3.24          &  -0.28         \\
N-MTLR                 & \cellcolor{green!25} 0.09           & \cellcolor{green!25} -0.11          & .0028          & -0.51          & \cellcolor{red!25} 5.32           &  0.42 \\
RPS                  & -0.03          & 0              & 0         & -0.97          & -1.28 &  -1.03         \\
RPS+Rank         & -0.03          & -0.06          & -.0017          & -1.19          & -2.51          & \cellcolor{green!25} -1.16         \\ \hline
\multicolumn{7}{c}{-Race}                                                                      \\ \hline
NLL                  & \cellcolor{green!25} 0.11           & \cellcolor{green!25} -0.14          & \cellcolor{green!25} -.0016          & \cellcolor{green!25} -1.95          & \cellcolor{green!25} -8.15          &  0.03          \\
DeepHit              & \cellcolor{green!25} 0.13           & -0.12          & 0          & \cellcolor{green!25} -2.6           & \cellcolor{green!25} -8.6           &  -0.63         \\
N-MTLR                 & \cellcolor{green!25} 0.21           & \cellcolor{green!25} -0.18          & \cellcolor{red!25} .0074          & -1.16          &  0.16           & -1.68         \\
RPS                  &  0              & 0.04           & -.0010          & \cellcolor{green!25} -1.38          & \cellcolor{green!25} -3.62          & \cellcolor{green!25} -0.88         \\
RPS+Rank         & 0.01           &  0.01           & .0002          & \cellcolor{green!25} -1.66          & \cellcolor{green!25} -4.46          & \cellcolor{green!25} -1.08         \\ \hline
\multicolumn{7}{c}{-Education}                                                                 \\ \hline
NLL                  & -0.01          & \cellcolor{green!25} -0.12          & \cellcolor{green!25} -0.0027          &  0.22           & \cellcolor{green!25} -4.09          & \cellcolor{green!25} -1.82         \\
DeepHit              & \cellcolor{green!25} 0.07           & \cellcolor{green!25} -0.27          & \cellcolor{green!25} -0.0029          & -0.69          &  -4.01          & \cellcolor{green!25} -2.12         \\
N-MTLR                 & \cellcolor{green!25} 0.03           & \cellcolor{green!25} -0.10          &  0.0018          & \cellcolor{red!25} 3.41           & 3.85           & -1.14         \\
RPS                  &  -0.12          & -0.01          & -0.0018          & \cellcolor{green!25} -1.71          &  -1.74          & \cellcolor{green!25} -2.93         \\
RPS+Rank         &  -0.16          & -0.05          & -0.0017          & \cellcolor{green!25} -1.98          &  -2.68          & \cellcolor{green!25} -3.1          \\ \hline
\end{tabular}
\caption{Change in performance with respect to Table \ref{tab:main-results}, where (-\textit{feature}) indicates a model trained without \textit{feature}. For clarity, cells which demonstrate a significant performance improvement (outside corresponding standard deviation range from Table \ref{tab:main-results}) are colored \colorbox{green!25}{green} while ones which show worse performance are \colorbox{red!25}{red}.}
\label{tab:abl-attr}
\end{table*}

Here, we consider the impact of the sensitive attributes in the training process. For this evaluation, we omit the selected attribute during training and then evaluate the resulting changes in both survival and fairness. Table \ref{tab:abl-attr} shows the change in performance with respect to the primary results due to the omission of the sensitive attributes. For clarity, we have highlighted the cells which demonstrate a \textit{significant change} in performance relative to initial results in Table \ref{tab:main-results}. We define significance as a change which is greater than the corresponding standard deviation reported in Table \ref{tab:main-results}.

We can see that the inclusion of \textit{sex} generally has minimal impact on the overall performance of the evaluated NDSMs, with its omission only yielding significant changes in a handful of categories. Meanwhile omitting \textit{race} attributes results in substantially better performance for nearly all models in terms of both survival and fairness. NLL, DeepHit, and N-MTLR each see a considerable improvement in discriminative performance with NLL and N-MTLR also experiencing further improvement in IBS. In terms of fairness, nearly every method improves for \textit{race} and \textit{sex}, with the RPS based methods improving in \textit{education} fairness as well. Finally, when omitting \textit{education} information, most models again improve with respect to fairness in the case of \textit{education}. However, unlike the \textit{race} attributes, there are fewer improvements in fairness across \textit{sex} and \textit{race}. 

These results indicate that is it generally acceptable to omit demographic features such as \textit{sex, race,} and \textit{education} from the training data as such features introduce additional bias without providing improvements in survival performance. In fact, these signals can introduce additional complexity which hinders the overall survival performance, particularly in the NLL, DeepHit, and N-MTLR methods.
It should also be noted that N-MTLR was the only method which did not have any fairness improvements across each of the three ablation experiments. This suggests that its bias is entirely learned implicitly from the non-demographic variables, which should be carefully considered when designing experiments for AD progression analysis pipelines.  

% compares these results to the performance when all features are included in the training process. We observe that by omitting these features, survival performance generally tends to worsen. This shows that, to some extent, there are revenant biomarkers in the sex and race features which are useful for survival predictions. However, we also note that removing sensitive attributes from the training process considerably improves the overall fairness of the model. This experiment suggests that there is some tradeoff between maximizing model fairness across sensitive populations and attaining the maximal discriminative performance.

\subsection{Permutation-Based Feature Importance Analysis}

Figure \ref{fig:importance} provides a representative example of the permutation feature importance analysis for the NLL model, averaged across three seeds. The remaining feature importance plots, along with relevant feature definitions, are provided in Appendix \ref{app:perm}.
The top five features were consistent across all models and appeared in the same order, with MEMORY, ORIENT, NACCAGE, JUDGEMENT, and CDRSUM ranking as the most influential predictors of AD progression. 
%“NACCAGE,” “ORIENT,” “MEMORY,” “NACCFAM\_1.0,” “NACCFAM\_0.0,” “COMPORT\_0.0,” and “CDRSUM” were consistently ranked among the top nine most important features. 
Overall, 13 features appeared within the top 20 importance rankings across all models, indicating substantial overlap in the predictors driving model performance despite differences in training objectives. This consistency suggests that model performance is primarily driven by stable clinical and cognitive indicators rather than loss-specific optimization effects.
% \begin{figure}[t]
% \label{fig:importance}

%     \centering
%     \includegraphics[width=1\linewidth]{figures/Feature_Importance.png}
%     \caption{Permutation-based feature importance analysis of study models (A) RPS Ranking, (B) RPS, (C) NLL, (D) DeepHit. }
%     \label{fig:importance}
% \end{figure}

\begin{figure}[t]
\label{fig:importance}

    \centering
    \includegraphics[width=1\linewidth]{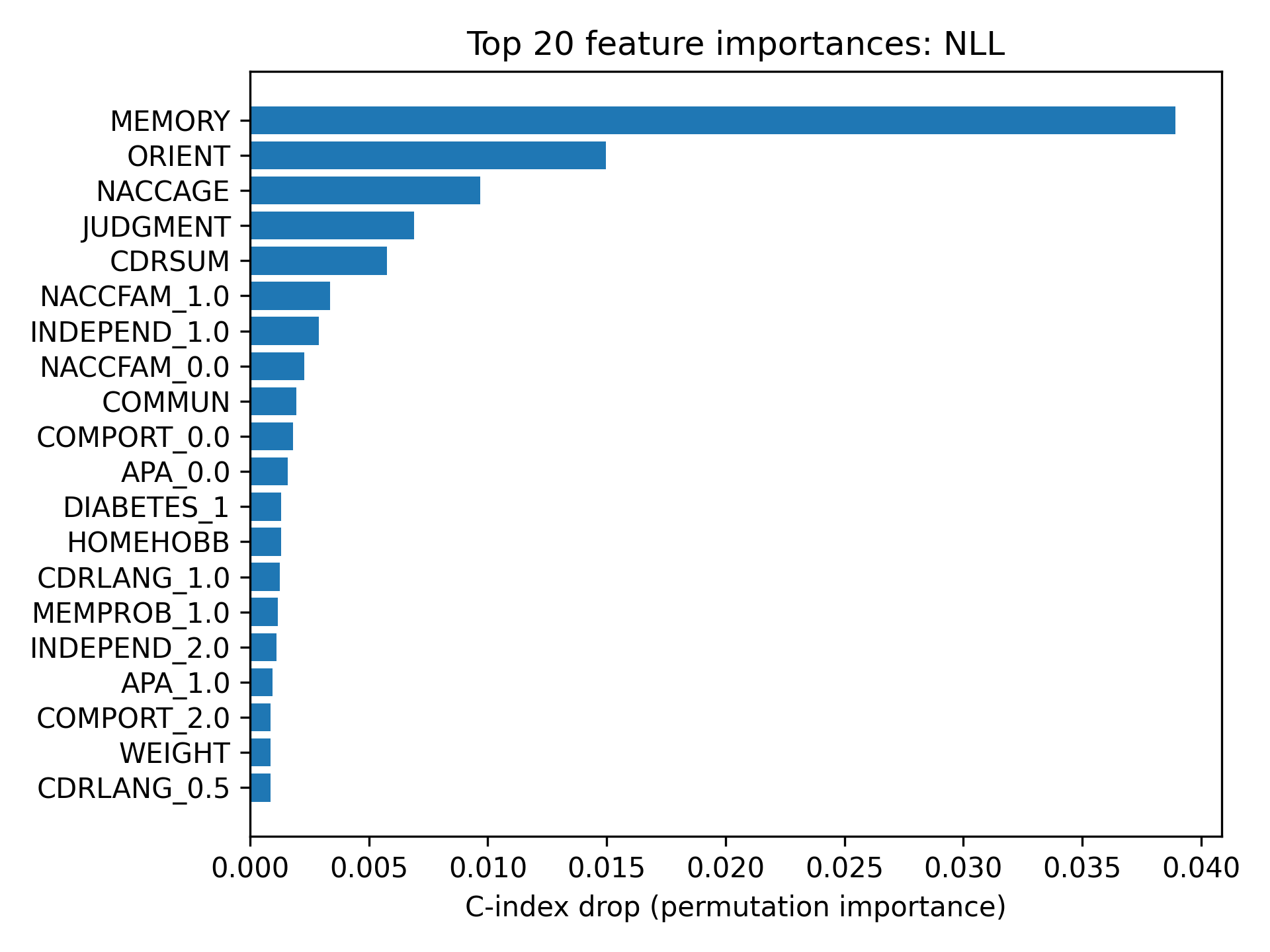}
    \caption{Permutation-based feature importance analysis of NLL model.}
    \label{fig:importance}
\end{figure}

\section{Discussion}

Our study finds that 
NDSMs are capable of capturing complex AD disease dynamics to attain strong discriminative performance while remaining calibrated with respect to the overall population. 
However, these methods often exhibit systematic fairness disparities across sensitive subgroups such as \textit{sex, ethnicity}, and \textit{education}. We find that these disparities can be partially mitigated at little to no cost to overall survival performance by omitting the relevant sensitive demographic features from the training data. While this does improve overall fairness, it does not fully mitigate bias, suggesting that NDSMs implicitly develop bias toward highly represented subgroups (e.g. white subjects). Finally, a permutation importance analysis revealed that the most important predictive features across all five NDSMs were \textit{identical}, indicating that such features are strong biomarkers for AD progression analysis, independent of the NDSM framework.

We additionally identify some important considerations of our approach. First, AD diagnosis is inherently noisy and subject to longitudinal instability, as the only definitive method to diagnose AD is post-mortem autopsy \cite{Gao2014-ta}. As a result, clinical diagnoses can fluctuate between visits due to evolving symptom presentation, changes in diagnostic criteria, or inter-rater variability. Although we use each subject's final recorded diagnosis to define disease conversion, this approach does not eliminate the risk of misclassifications. Diagnostic noise directly impacts survival time estimation and  complicates fairness evaluation, as fairness metrics implicitly assume that the model's target labels are correct. In the absence of a definitive ground truth for AD progression, observed performance differences across demographic groups may therefore reflect diagnostic variability rather than true differences in disease dynamics.

Second, our analysis relies on standard survival model assumptions regarding censoring that may not indeed hold true. In particular, the assumption of independent censoring may be violated if patients who drop out of the study differ systematically from those who remained in data collection. Differential loss to follow-up may be associated with disease severity, socioeconomic factors, or access to care, and these factors may themselves vary across different demographic groups, potentially biasing our results.

Finally, the use of NACC data introduces potential sample bias, as participants in this data collection effort are typically recruited through specialized AD research centers and receive frequent structured follow-up assessments \cite{Chan2025-pz}. As a result, this cohort may not be representative of the broader populations of individuals with AD, particularly those with limited access to specialty care. Consequently, this study's model performance and fairness conclusions may not generalize well to under-resourced clinical settings.

    % Finally, cognitive status indicators such as “DEMENTED” and “NORMCOG” are highly informative and consistently ranked among the most important features, but their inclusion may complicate interpretation through partial encoding of downstream clinical outcomes. We chose to retain these variables in our present analysis because of the fact that they are clinically available at baseline and are commonly used in prior work, such as that of Sharma et al \cite{Sharma2021-ll}. Nevertheless, their presence may inflate apparent predictive performance and obscure the contributions of other covariates.

    % Finally, the survival models evaluated in this work are sensitive to design choices related to time discretization. The selection of time-bin granularity can substantially influence model calibration, risk estimation, and downstream fairness metrics, and different discretization schemes may lead to different conclusions regarding model performance across demographic groups.

\section{Conclusion}

This work introduces a comprehensive evaluation framework for nonparametric deep survival models in Alzheimer’s disease progression analysis, moving beyond discrimination alone to jointly assess calibration, fairness, and interpretability. 
Within this framework, we propose two complementary fairness metrics—\textit{Time-dependent Concordance Impurity} and \textit{Kaplan–Meier Fairness}—designed to quantify subgroup disparities in predicted survival dynamics rather than static risk scores.
% This work establishes a framework for evaluating holistic performance of Nonparametric Deep Survival Models for Alzheimer's disease progression analysis by analyzing model discrimination, calibration, and fairness. To this end, we propose two fairness metrics: \textbf{Time-dependent Concordance Impurity} and \textbf{Kaplan-Meier Fairness} to evaluate the systematic disparities learned by the model across various subgroups.
% We also examine feature importance for AD progression through a rigorous permutation analysis and demonstrate that there exist a set of stable biomarkers which carry strong predictive signals across five NDSMs.
Our analysis further provides a systematic investigation of feature stability via permutation-based importance, revealing a core set of biomarkers that consistently drive predictions across five distinct NDSMs, suggesting the presence of robust, model-agnostic signals of disease progression.
Collectively, these results highlight the importance of multidimensional evaluation for determining clinical readiness and provide a foundation for developing more reliable and equitable deep survival models.
% We hope this framework can act as a foundation toward developing stronger and more reliable NDSMs.

% \section{Acknowledgments}

% Identification of funding sources and other support, and thanks to
% individuals and groups that assisted in the research and the
% preparation of the work should be included in an acknowledgment
% section, which is placed just before the reference section in your
% document.

% This section has a special environment:
% \begin{verbatim}
%   \begin{acks}
%   ...
%   \end{acks}
% \end{verbatim}
% so that the information contained therein can be more easily collected
% during the article metadata extraction phase, and to ensure
% consistency in the spelling of the section heading.

% Authors should not prepare this section as a numbered or unnumbered {\verb|\section|}; please use the ``{\verb|acks|}'' environment.

% \section{Appendices}

% If your work needs an appendix, add it before the
% ``\verb|\end{document}|'' command at the conclusion of your source
% document.

% Start the appendix with the ``\verb|appendix|'' command:
% \begin{verbatim}
%   \appendix
% \end{verbatim}
% and note that in the appendix, sections are lettered, not
% numbered. This document has two appendices, demonstrating the section
% and subsection identification method.


\bibliographystyle{ACM-Reference-Format}
\bibliography{main}

@String(AAAI = {AAAI})

@misc{kingma2017adammethodstochasticoptimization,
      title={Adam: A Method for Stochastic Optimization}, 
      author={Diederik P. Kingma and Jimmy Ba},
      year={2017},
      eprint={1412.6980},
      archivePrefix={arXiv},
      primaryClass={cs.LG},
      url={https://arxiv.org/abs/1412.6980}, 
}

@article{Tang_Zhang_Li_2025, title={From Representation Space to Prognostic Insights: Whole Slide Image Generation with Hierarchical Diffusion Model for Survival Prediction}, volume={39}, url={https://ojs.aaai.org/index.php/AAAI/article/view/32788}, DOI={10.1609/aaai.v39i7.32788}, abstractNote={Deep learning has significantly enhanced survival prediction using whole slide images (WSIs) by adopting a two-stage learning paradigm: WSI preparation and patient-level prediction. While existing research generally concentrates on developing advanced patient-level prediction modules, the critical importance of WSI preparation has been largely overlooked. In practice, WSI preparation is influenced by numerous factors, including tissue heterogeneity, sampling strategies, and technical considerations. These uncontrollable external factors incur variability in the number of WSIs among patients, introducing significant bias and resulting in inferior performance for patients with few WSIs. To address this challenge, we propose a novel approach named WSI-Diffusion. Unlike existing WSI generation models that produce augmented versions of input WSIs, our method generates entirely new WSIs in representation space to serve as complementary data. WSIDiffusion employs a two-stage hierarchical diffusion process. Two novel modules, WSI-level and patch-level Diffusers are designed to capture complex correlations between WSIs and patches. The generated WSIs are integrated as supplementary data, and a light patient-level prediction module is then trained for survival prediction. Experimental results across five datasets demonstrate the superiority of our proposal.}, number={7}, journal={Proceedings of the AAAI Conference on Artificial Intelligence}, author={Tang, Zhihao and Zhang, Xi and Li, Chaozhuo}, year={2025}, month={Apr.}, pages={7329-7337} }

@article{sheng2025learning,
  title={Learning Survival Distributions with the Asymmetric Laplace Distribution},
  author={Sheng, Deming and Henao, Ricardo},
  journal={arXiv preprint arXiv:2505.03712},
  year={2025}
}

@ARTICLE{lognorm-mle,
  title     = "A new analytical potential energy surface for the adsorption
               system {CO/Cu(100})",
  author    = "Marquardt, Roberto and Cuvelier, Fr{\'e}d{\'e}ric and Olsen,
               Roar A and Baerends, Evert Jan and Tremblay, Jean Christophe and
               Saalfrank, Peter",
  journal   = "J. Chem. Phys.",
  publisher = "AIP Publishing",
  volume    =  132,
  number    =  7,
  pages     = "074108",
  month     =  feb,
  year      =  2010,
  language  = "en"
}

@misc{deepweisurv,
      title={Estimation of conditional mixture Weibull distribution with right-censored data using neural network for time-to-event analysis}, 
      author={Achraf Bennis and Sandrine Mouysset and Mathieu Serrurier},
      year={2020},
      eprint={2002.09358},
      archivePrefix={arXiv},
      primaryClass={stat.ME},
      url={https://arxiv.org/abs/2002.09358}, 
}

@ARTICLE{XCAL,
  author={Chapfuwa, Paidamoyo and Tao, Chenyang and Li, Chunyuan and Khan, Irfan and Chandross, Karen J. and Pencina, Michael J. and Carin, Lawrence and Henao, Ricardo},
  journal={IEEE Transactions on Neural Networks and Learning Systems}, 
  title={Calibration and Uncertainty in Neural Time-to-Event Modeling}, 
  year={2023},
  volume={34},
  number={4},
  pages={1666-1680},
  doi={10.1109/TNNLS.2020.3029631}}

@article{Kaplan01061958,
author = {E. L. Kaplan and Paul Meier},
title = {Nonparametric Estimation from Incomplete Observations},
journal = {Journal of the American Statistical Association},
volume = {53},
number = {282},
pages = {457--481},
year = {1958},
publisher = {Taylor \& Francis},
doi = {10.1080/01621459.1958.10501452},
}

@inproceedings{qi2024toward,
 author =        {Qi, Shi-ang and Yu, Yakun and Greiner, Russell},
 booktitle =     {Advances in Neural Information Processing Systems},
 pages =         {86180--86225},
 publisher =     {Curran Associates, Inc.},
 title =         {Toward Conditional Distribution Calibration in Survival Prediction},
 url =           {https://proceedings.neurips.cc/paper_files/paper/2024/file/9c8df8de46c1a1b39b30b9f74be69c02-Paper-Conference.pdf},
 volume =        {37},
 year =          {2024}
}

@InProceedings{pmlr-v235-qi24a,
  title =        {Conformalized Survival Distributions: A Generic Post-Process to Increase Calibration},
  author =       {Qi, Shi-Ang and Yu, Yakun and Greiner, Russell},
  booktitle = 	 {Proceedings of the 41st International Conference on Machine Learning},
  pages = 	     {41303--41339},
  year = 	       {2024},
  volume = 	     {235},
  series = 	     {Proceedings of Machine Learning Research},
  month = 	     {21--27 Jul},
  publisher =    {PMLR},
  url = 	       {https://proceedings.mlr.press/v235/qi24a.html},
}

@inproceedings{
lee2024toward,
title={Toward a Well-Calibrated Discrimination via Survival Outcome-Aware Contrastive Learning},
author={Dongjoon Lee and Hyeryn Park and Changhee Lee},
booktitle={The Thirty-eighth Annual Conference on Neural Information Processing Systems},
year={2024},
url={https://openreview.net/forum?id=UVjuYBSbCN}
}

@article{KRZYZINSKI2023110234,
title = {SurvSHAP(t): Time-dependent explanations of machine learning survival models},
journal = {Knowledge-Based Systems},
volume = {262},
pages = {110234},
year = {2023},
issn = {0950-7051},
doi = {https://doi.org/10.1016/j.knosys.2022.110234},
url = {https://www.sciencedirect.com/science/article/pii/S0950705122013302},
author = {Mateusz Krzyziński and Mikołaj Spytek and Hubert Baniecki and Przemysław Biecek}
}

@article{survshap,
    title = {SurvSHAP(t): Time-dependent explanations of machine learning survival models},
    author = {Mateusz Krzyziński and Mikołaj Spytek and Hubert Baniecki and Przemysław Biecek},
    journal = {Knowledge-Based Systems},
    volume = {262},
    pages = {110234},
    year = {2023}
}

@article{rps11, 
title={Estimating Calibrated Individualized Survival Curves with Deep Learning}, volume={35}, 
url={https://ojs.aaai.org/index.php/AAAI/article/view/16098}, 
DOI={10.1609/aaai.v35i1.16098}, 
abstractNote={In survival analysis, deep learning approaches have been proposed for estimating an individual’s probability of survival over some time horizon. Such approaches can capture complex non-linear relationships, without relying on restrictive assumptions regarding the relationship between an individual’s characteristics and their underlying survival process. To date, however, these methods have focused primarily on optimizing discriminative performance and have ignored model calibration. Well-calibrated survival curves present realistic and meaningful probabilistic estimates of the true underlying survival process for an individual. However, due to the lack of ground-truth regarding the underlying stochastic process of survival for an individual, optimizing and measuring calibration in survival analysis is an inherently difficult task. In this work, we i) highlight the shortcomings of existing approaches in terms of calibration and ii) propose a new training scheme for optimizing deep survival analysis models that maximizes discriminative performance, subject to good calibration. Compared to state-of-the-art approaches across two publicly available datasets, our proposed training scheme leads to significant improvements in calibration, while maintaining good discriminative performance.}, 
number={1}, 
journal={Proceedings of the AAAI Conference on Artificial Intelligence}, 
author={Kamran, Fahad and Wiens, Jenna}, 
year={2021}, 
month={May}, 
pages={240-248} }

@misc{MTLR,
      title={Deep Neural Networks for Survival Analysis Based on a Multi-Task Framework}, 
      author={Stephane Fotso},
      year={2018},
      eprint={1801.05512},
      archivePrefix={arXiv},
      primaryClass={stat.ML},
      url={https://arxiv.org/abs/1801.05512}, 
}

@Misc{pysurvival_cite,
  author = {Stephane Fotso and others},
  title = {{PySurvival}: Open source package for Survival Analysis modeling},
  year = {2019--},
  url = "https://www.pysurvival.io/"
}

@article{Antolini2005ATD,
  title={A time‐dependent discrimination index for survival data},
  author={Laura Antolini and Patrizia Boracchi and Elia M Biganzoli},
  journal={Statistics in Medicine},
  year={2005},
  volume={24},
  url={https://api.semanticscholar.org/CorpusID:25663825}
}

@misc{pycox,
      title={Continuous and Discrete-Time Survival Prediction with Neural Networks}, 
      author={Håvard Kvamme and Ørnulf Borgan},
      year={2019},
      eprint={1910.06724},
      archivePrefix={arXiv},
      primaryClass={stat.ML},
      url={https://arxiv.org/abs/1910.06724}, 
}

@article{clark2003survival,
  title={Survival analysis part I: basic concepts and first analyses},
  author={Clark, Taane G and Bradburn, Michael J and Love, Sharon B and Altman, Douglas G},
  journal={British journal of cancer},
  volume={89},
  number={2},
  pages={232--238},
  year={2003},
  publisher={Nature Publishing Group}
}

@article{wiegrebe2024deep,
  title={Deep learning for survival analysis: a review},
  author={Wiegrebe, Simon and Kopper, Philipp and Sonabend, Raphael and Bischl, Bernd and Bender, Andreas},
  journal={Artificial Intelligence Review},
  volume={57},
  number={3},
  pages={65},
  year={2024},
  publisher={Springer}
}

@misc{thrasher2024tessltimeeventawareself,
      title={TE-SSL: Time and Event-aware Self Supervised Learning for Alzheimer's Disease Progression Analysis}, 
      author={Jacob Thrasher and Alina Devkota and Ahmed Tafti and Binod Bhattarai and Prashnna Gyawali},
      year={2024},
      eprint={2407.06852},
      archivePrefix={arXiv},
      primaryClass={cs.CV},
      url={https://arxiv.org/abs/2407.06852}, 
}

@misc{saeed2024survrnclearningorderedrepresentations,
      title={SurvRNC: Learning Ordered Representations for Survival Prediction using Rank-N-Contrast}, 
      author={Numan Saeed and Muhammad Ridzuan and Fadillah Adamsyah Maani and Hussain Alasmawi and Karthik Nandakumar and Mohammad Yaqub},
      year={2024},
      eprint={2403.10603},
      archivePrefix={arXiv},
      primaryClass={cs.CV},
      url={https://arxiv.org/abs/2403.10603}, 
}

@misc{xu2025distilledpromptlearningincomplete,
      title={Distilled Prompt Learning for Incomplete Multimodal Survival Prediction}, 
      author={Yingxue Xu and Fengtao Zhou and Chenyu Zhao and Yihui Wang and Can Yang and Hao Chen},
      year={2025},
      eprint={2503.01653},
      archivePrefix={arXiv},
      primaryClass={cs.LG},
      url={https://arxiv.org/abs/2503.01653}, 
}

@misc{huo2025timetoeventpretraining3dmedical,
      title={Time-to-Event Pretraining for 3D Medical Imaging}, 
      author={Zepeng Huo and Jason Alan Fries and Alejandro Lozano and Jeya Maria Jose Valanarasu and Ethan Steinberg and Louis Blankemeier and Akshay S. Chaudhari and Curtis Langlotz and Nigam H. Shah},
      year={2025},
      eprint={2411.09361},
      archivePrefix={arXiv},
      primaryClass={cs.CV},
      url={https://arxiv.org/abs/2411.09361}, 
}

@misc{zhang2022longitudinalfairnesscensorship,
      title={Longitudinal Fairness with Censorship}, 
      author={Wenbin Zhang and Jeremy C. Weiss},
      year={2022},
      eprint={2203.16024},
      archivePrefix={arXiv},
      primaryClass={cs.LG},
      url={https://arxiv.org/abs/2203.16024}, 
}

@inproceedings{deephit,
  title={Deephit: A deep learning approach to survival analysis with competing risks},
  author={Lee, Changhee and Zame, William and Yoon, Jinsung and Van Der Schaar, Mihaela},
  booktitle={Proceedings of the AAAI conference on artificial intelligence},
  volume={32},
  number={1},
  year={2018}
}

@article{rps, 
title={Estimating Calibrated Individualized Survival Curves with Deep Learning}, volume={35}, 
url={https://ojs.aaai.org/index.php/AAAI/article/view/16098}, 
DOI={10.1609/aaai.v35i1.16098}, 
abstractNote={In survival analysis, deep learning approaches have been proposed for estimating an individual’s probability of survival over some time horizon. Such approaches can capture complex non-linear relationships, without relying on restrictive assumptions regarding the relationship between an individual’s characteristics and their underlying survival process. To date, however, these methods have focused primarily on optimizing discriminative performance and have ignored model calibration. Well-calibrated survival curves present realistic and meaningful probabilistic estimates of the true underlying survival process for an individual. However, due to the lack of ground-truth regarding the underlying stochastic process of survival for an individual, optimizing and measuring calibration in survival analysis is an inherently difficult task. In this work, we i) highlight the shortcomings of existing approaches in terms of calibration and ii) propose a new training scheme for optimizing deep survival analysis models that maximizes discriminative performance, subject to good calibration. Compared to state-of-the-art approaches across two publicly available datasets, our proposed training scheme leads to significant improvements in calibration, while maintaining good discriminative performance.}, 
number={1}, 
journal={Proceedings of the AAAI Conference on Artificial Intelligence}, 
author={Kamran, Fahad and Wiens, Jenna}, 
year={2021}, 
month={May}, 
pages={240-248} }

@article{Katzman_2018,
   title={DeepSurv: personalized treatment recommender system using a Cox proportional hazards deep neural network},
   volume={18},
   ISSN={1471-2288},
   url={http://dx.doi.org/10.1186/s12874-018-0482-1},
   DOI={10.1186/s12874-018-0482-1},
   number={1},
   journal={BMC Medical Research Methodology},
   publisher={Springer Science and Business Media LLC},
   author={Katzman, Jared L. and Shaham, Uri and Cloninger, Alexander and Bates, Jonathan and Jiang, Tingting and Kluger, Yuval},
   year={2018},
   month=feb }

@inproceedings{zhang-fairness,
author = {Zhang, Wenbin and Hernandez-Boussard, Tina and Weiss, Jeremy},
title = {Censored fairness through awareness},
year = {2023},
isbn = {978-1-57735-880-0},
publisher = {AAAI Press},
url = {https://doi.org/10.1609/aaai.v37i12.26708},
doi = {10.1609/aaai.v37i12.26708},
booktitle = {Proceedings of the Thirty-Seventh AAAI Conference on Artificial Intelligence and Thirty-Fifth Conference on Innovative Applications of Artificial Intelligence and Thirteenth Symposium on Educational Advances in Artificial Intelligence},
articleno = {1639},
numpages = {9},
series = {AAAI'23/IAAI'23/EAAI'23}
}

@article{Liu2023ATPA,
  title={A translational perspective towards clinical AI fairness},
  author={Mingxuan Liu and Yilin Ning and Salinelat Teixayavong and M. Mertens and Jie Xu and D. Ting and L. T. Cheng and J. Ong and Zhen Ling Teo and Ting Fang Tan and Ravi Chandran Narrendar and Fei Wang and L. Celi and M. Ong and Nan Liu},
  journal={NPJ Digital Medicine},
  year={2023},
  volume={6},
  url={https://api.semanticscholar.org/CorpusId:261883775}
}

@article{Yang2024DemographicBOA,
  title={Demographic Bias of Expert-Level Vision-Language Foundation Models in Medical Imaging},
  author={Yuzhe Yang and Yujia Liu and Xin Liu and Avanti V Gulhane and Domenico Mastrodicasa and Wei Wu and E. J. Wang and Dushyant W. Sahani and Shwetak N. Patel},
  journal={Science Advances},
  year={2024},
  volume={11},
  url={https://api.semanticscholar.org/CorpusId:267782475}
}

@ARTICLE{Goktas2025,
  title     = "Shaping the future of healthcare: Ethical clinical challenges
               and pathways to trustworthy {AI}",
  author    = "Goktas, Polat and Grzybowski, Andrzej",
  journal   = "J. Clin. Med.",
  publisher = "MDPI AG",
  volume    =  14,
  number    =  5,
  pages     = "1605",
  month     =  feb,
  year      =  2025,
  copyright = "https://creativecommons.org/licenses/by/4.0/",
  language  = "en"
}

@ARTICLE{Huh2020,
  title     = "Survival analysis of patients with Alzheimer's disease: A study
               based on data from the Korean National Health Insurance
               Services' Senior Cohort database",
  author    = "Huh, Tae Ho and Yoon, Jong Lull and Cho, Jung Jin and Kim, Mee
               Young and Ju, Young Soo",
  journal   = "Korean J. Fam. Med.",
  publisher = "The Korean Academy of Family Medicine",
  volume    =  41,
  number    =  4,
  pages     = "214--221",
  month     =  jul,
  year      =  2020,
  language  = "en"
}

@ARTICLE{Abuhantash2025,
  title    = "Alzheimer's disease risk prediction using machine learning for
              survival analysis with a comorbidity-based approach",
  author   = "Abuhantash, Ferial and Welsch, Roy and Finkelstein, Stan and
              AlShehhi, Aamna",
  journal  = "Scientific Reports",
  volume   =  15,
  number   =  1,
  pages    = "28723",
  month    =  aug,
  year     =  2025
}

@article{AD-time-to-event,
author = {Sharma, Rahul and Anand, Harsh and Badr, Youakim and Qiu, Robin G.},
title = {Time-to-event prediction using survival analysis methods for Alzheimer's disease progression},
journal = {Alzheimer's \& Dementia: Translational Research \& Clinical Interventions},
volume = {7},
number = {1},
pages = {e12229},
doi = {https://doi.org/10.1002/trc2.12229},
url = {https://alz-journals.onlinelibrary.wiley.com/doi/abs/10.1002/trc2.12229},
eprint = {https://alz-journals.onlinelibrary.wiley.com/doi/pdf/10.1002/trc2.12229},
year = {2021}
}

@ARTICLE{Kukull2025,
  title     = "The National Alzheimer's Coordinating Center ({NACC}) 1999-2025:
               Personal history and recollections",
  author    = "Kukull, Walter A",
  journal   = "Alzheimers. Dement.",
  publisher = "Wiley",
  volume    =  21,
  number    =  10,
  pages     = "e70836",
  month     =  oct,
  year      =  2025,
  copyright = "http://creativecommons.org/licenses/by-nc-nd/4.0/",
  language  = "en"
}

@article{AlzheimersAssociation2024FactsFigures,
  title   = {2024 Alzheimer's disease facts and figures},
  author  = {{Alzheimer's Association}},
  journal = {Alzheimer's \& Dementia},
  volume  = {20},
  number  = {5},
  pages   = {3708--3821},
  year    = {2024},
  month   = {may},
  doi     = {10.1002/alz.13809},
  pmid    = {38689398},
}

@ARTICLE{Duara2022-bt,
  title     = "Heterogeneity in Alzheimer's disease diagnosis and progression
               rates: Implications for therapeutic trials",
  author    = "Duara, Ranjan and Barker, Warren",
  journal   = "Neurotherapeutics",
  publisher = "Elsevier BV",
  volume    =  19,
  number    =  1,
  pages     = "8--25",
  month     =  jan,
  year      =  2022,
  copyright = "http://www.elsevier.com/open-access/userlicense/1.0/",
  language  = "en"
}

@ARTICLE{Sperling2013-bt,
  title     = "Preclinical Alzheimer disease---the challenges ahead",
  author    = "Sperling, Reisa A and Karlawish, Jason and Johnson, Keith A",
  journal   = "Nat. Rev. Neurol.",
  publisher = "Springer Science and Business Media LLC",
  volume    =  9,
  number    =  1,
  pages     = "54--58",
  month     =  jan,
  year      =  2013,
  language  = "en"
}

@ARTICLE{Nakashima2025-yy,
  title     = "Therapeutic time window of disease-modifying therapy for early
               Alzheimer's disease",
  author    = "Nakashima, Saki and Sato, Kenichiro and Niimi, Yoshiki and
               Ihara, Ryoko and Suzuki, Kazushi and Iwata, Atsushi and Toda,
               Tatsushi and Iwatsubo, Takeshi and {for Alzheimer's Disease
               Neuroimaging Initiative}",
  journal   = "Alzheimers Dement. (N. Y.)",
  publisher = "Wiley",
  volume    =  11,
  number    =  2,
  pages     = "e70102",
  month     =  apr,
  year      =  2025,
  copyright = "http://creativecommons.org/licenses/by-nc/4.0/",
  language  = "en"
}

@article {ANewVectorPartitionoftheProbabilityScore,
      author = "Allan H.  Murphy",
      title = "A New Vector Partition of the Probability Score",
      journal = "Journal of Applied Meteorology and Climatology",
      year = "1973",
      publisher = "American Meteorological Society",
      address = "Boston MA, USA",
      volume = "12",
      number = "4",
      doi = "10.1175/1520-0450(1973)012<0595:ANVPOT>2.0.CO;2",
      pages=      "595 - 600",
      url = "https://journals.ametsoc.org/view/journals/apme/12/4/1520-0450_1973_012_0595_anvpot_2_0_co_2.xml"
}

@article{Hosmer01011980,
author = {David W. Hosmer and Stanley Lemesbow},
title = {Goodness of fit tests for the multiple logistic regression model},
journal = {Communications in Statistics - Theory and Methods},
volume = {9},
number = {10},
pages = {1043--1069},
year = {1980},
publisher = {Taylor \& Francis},
doi = {10.1080/03610928008827941},


URL = { 
    
    
        https://www.tandfonline.com/doi/abs/10.1080/03610928008827941
    

},
eprint = { 
    
    
        https://www.tandfonline.com/doi/pdf/10.1080/03610928008827941
    

}}

@ARTICLE{Mayeda2016-qb,
  title     = "Inequalities in dementia incidence between six racial and ethnic
               groups over 14 years",
  author    = "Mayeda, Elizabeth Rose and Glymour, M Maria and Quesenberry,
               Charles P and Whitmer, Rachel A",
  journal   = "Alzheimers. Dement.",
  publisher = "Wiley",
  volume    =  12,
  number    =  3,
  pages     = "216--224",
  month     =  mar,
  year      =  2016,
  copyright = "http://onlinelibrary.wiley.com/termsAndConditions\#vor",
  language  = "en"
}

@ARTICLE{Lewis2023-xb,
  title     = "Association between socioeconomic factors, race, and use of a
               specialty memory clinic",
  author    = "Lewis, Abigail and Gupta, Aditi and Oh, Inez and Schindler,
               Suzanne E and Ghoshal, Nupur and Abrams, Zachary and Foraker,
               Randi and Snider, Barbara Joy and Morris, John C and
               Balls-Berry, Joyce and Gupta, Mahendra and Payne, Philip R O and
               Lai, Albert M",
  journal   = "Neurology",
  publisher = "Ovid Technologies (Wolters Kluwer Health)",
  volume    =  101,
  number    =  14,
  pages     = "e1424--e1433",
  month     =  oct,
  year      =  2023,
  language  = "en"
}

@ARTICLE{Sharma2021-ll,
  title     = "Time-to-event prediction using survival analysis methods for
               Alzheimer's disease progression",
  author    = "Sharma, Rahul and Anand, Harsh and Badr, Youakim and Qiu, Robin
               G",
  journal   = "Alzheimers Dement. (N. Y.)",
  publisher = "Wiley",
  volume    =  7,
  number    =  1,
  pages     = "e12229",
  month     =  dec,
  year      =  2021,
  copyright = "http://creativecommons.org/licenses/by-nc-nd/4.0/",
  language  = "en"
}

@ARTICLE{Gao2014-ta,
  title     = "Mild cognitive impairment, incidence, progression, and
               reversion: findings from a community-based cohort of elderly
               African Americans",
  author    = "Gao, Sujuan and Unverzagt, Frederick W and Hall, Kathleen S and
               Lane, Kathleen A and Murrell, Jill R and Hake, Ann M and
               Smith-Gamble, Valerie and Hendrie, Hugh C",
  journal   = "Am. J. Geriatr. Psychiatry",
  publisher = "Elsevier BV",
  volume    =  22,
  number    =  7,
  pages     = "670--681",
  month     =  jul,
  year      =  2014,
  language  = "en"
}

@ARTICLE{Chan2025-pz,
  title     = "{NACC} data: Who is represented over time and across centers,
               and implications for generalizability",
  author    = "Chan, Kwun C G and Xia, Fan and Kukull, Walter A",
  journal   = "Alzheimers. Dement.",
  publisher = "Wiley",
  volume    =  21,
  number    =  9,
  pages     = "e70657",
  month     =  sep,
  year      =  2025,
  copyright = "http://creativecommons.org/licenses/by/4.0/",
  language  = "en"
}

\appendix

\section{Selected features}
\label{app:feat}

\begin{figure}[t]
  \centering
   \includegraphics[width=\linewidth]{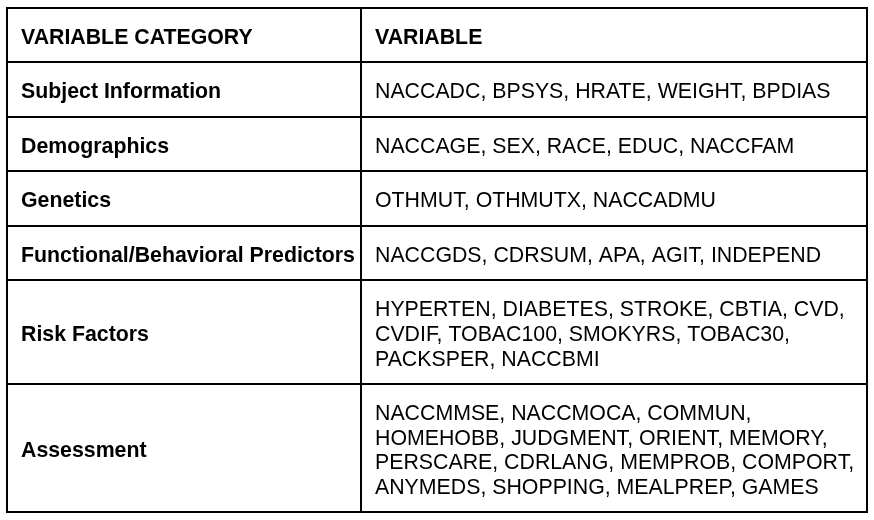}

   \caption{Selected features from the NACC dataset.}
   \label{fig:features}
\end{figure}

 Figure \ref{fig:features} provides an overview of the NACC features used in our analysis. We categorize features into six groups: subject visit information, demographics, genetics, functional/behavior predictors, risk factors, and assessment features.
 Importantly, OTHMUT, OTHMUTX, and NACCMMSE were dropped during imputation due to excessive missingness (>30\%).

\begin{figure*}[ht]
     \centering
     \begin{subfigure}[t]{.45\textwidth}
         \centering
         \includegraphics[width=\textwidth]{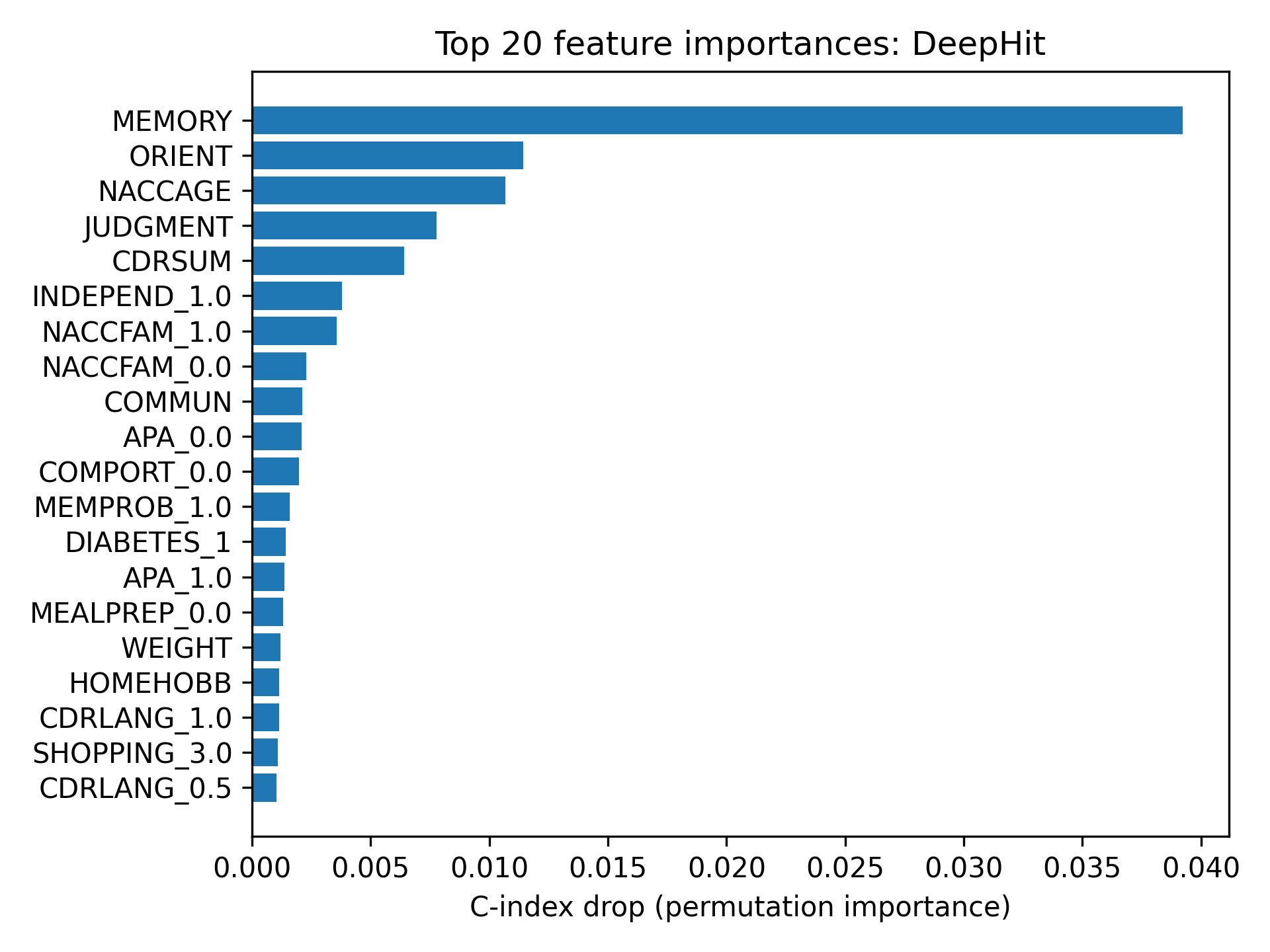}
     \end{subfigure}
     \hfill
     \begin{subfigure}[t]{.45\textwidth}
         \centering
         \includegraphics[width=\textwidth]{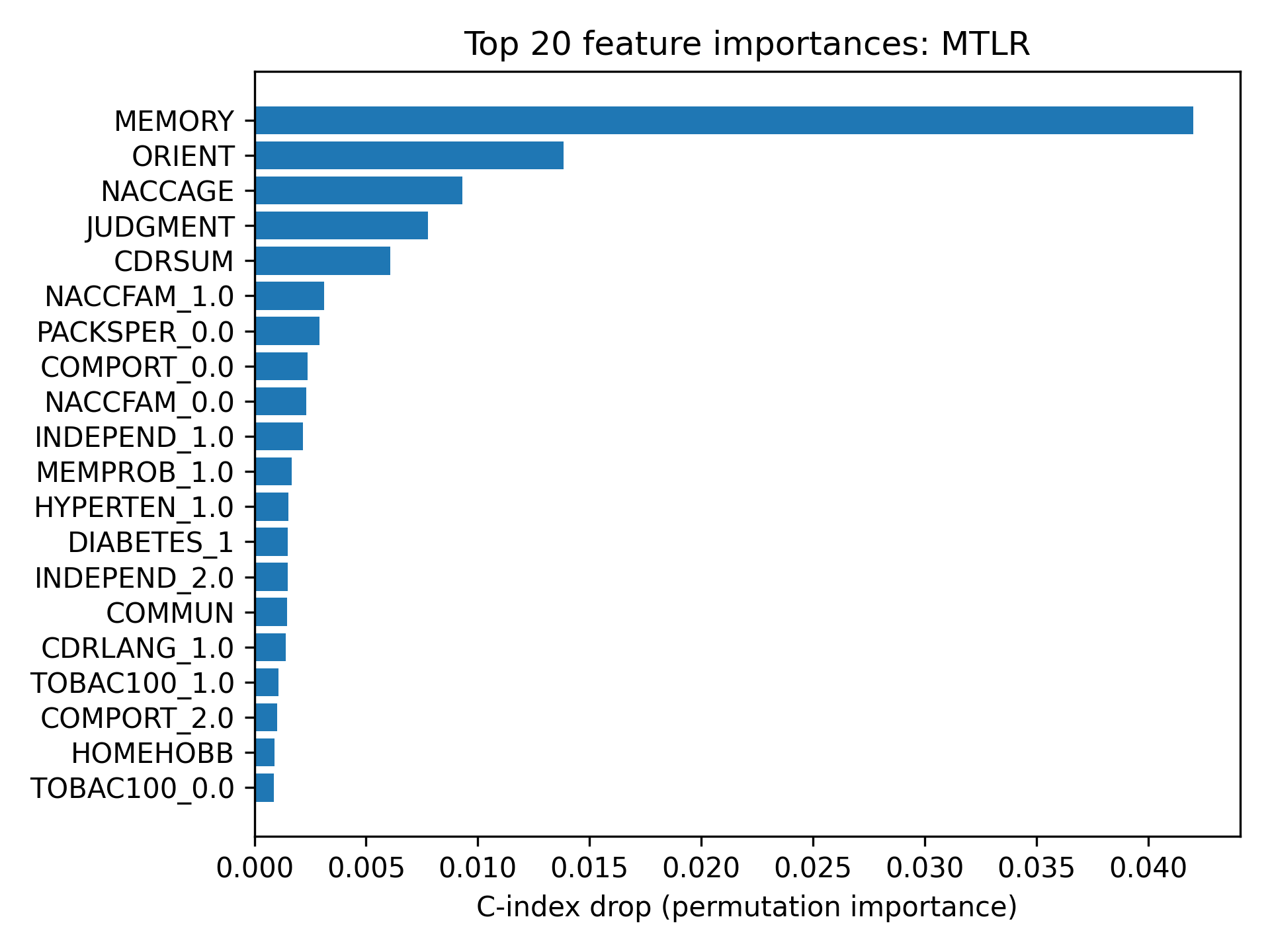}
     \end{subfigure}
     \hfill
     \begin{subfigure}[t]{.45\textwidth}
         \centering
         \includegraphics[width=\textwidth]{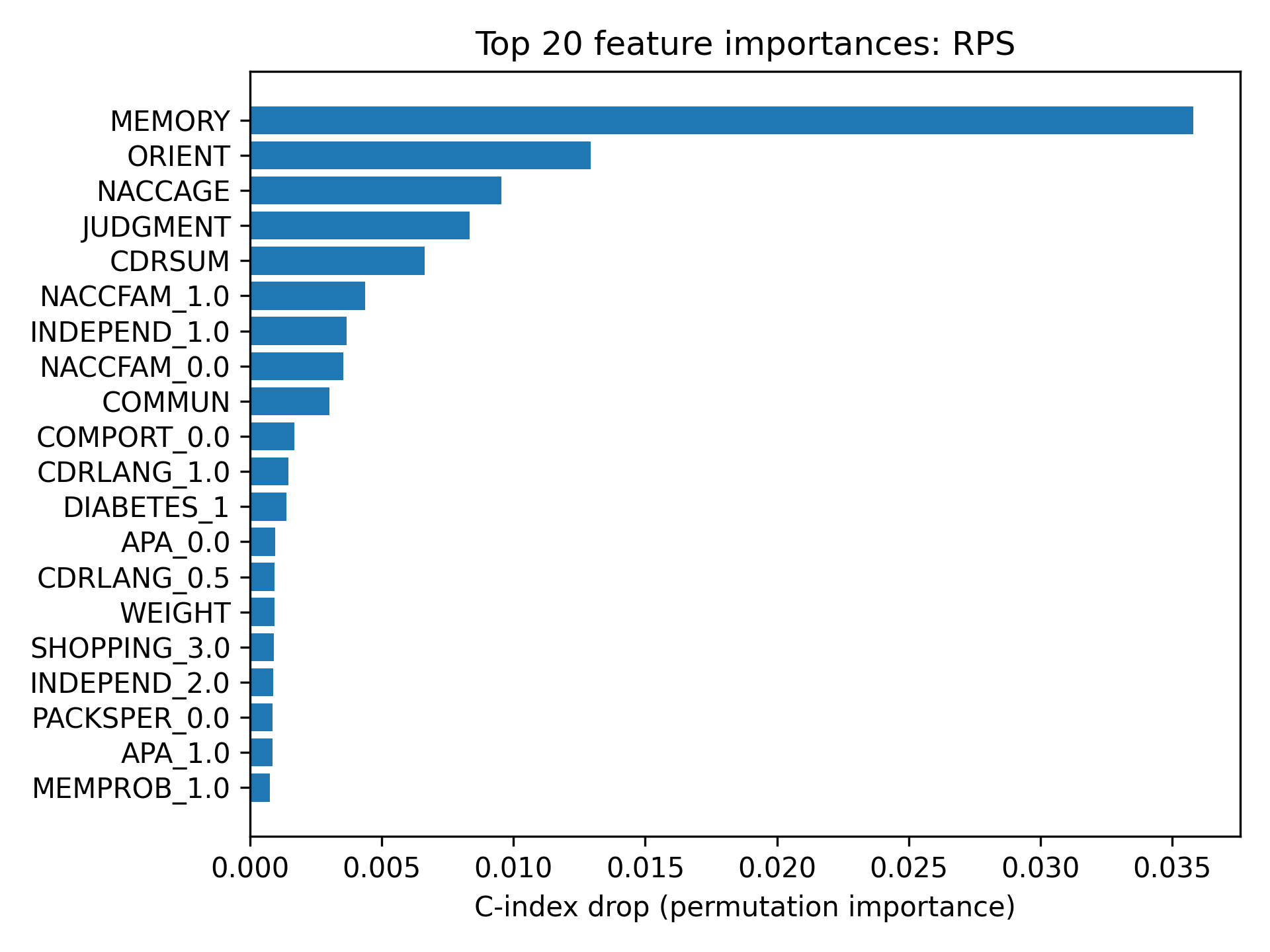}
     \end{subfigure}
     \hfill
     \begin{subfigure}[t]{.45\textwidth}
         \centering
         \includegraphics[width=\textwidth]{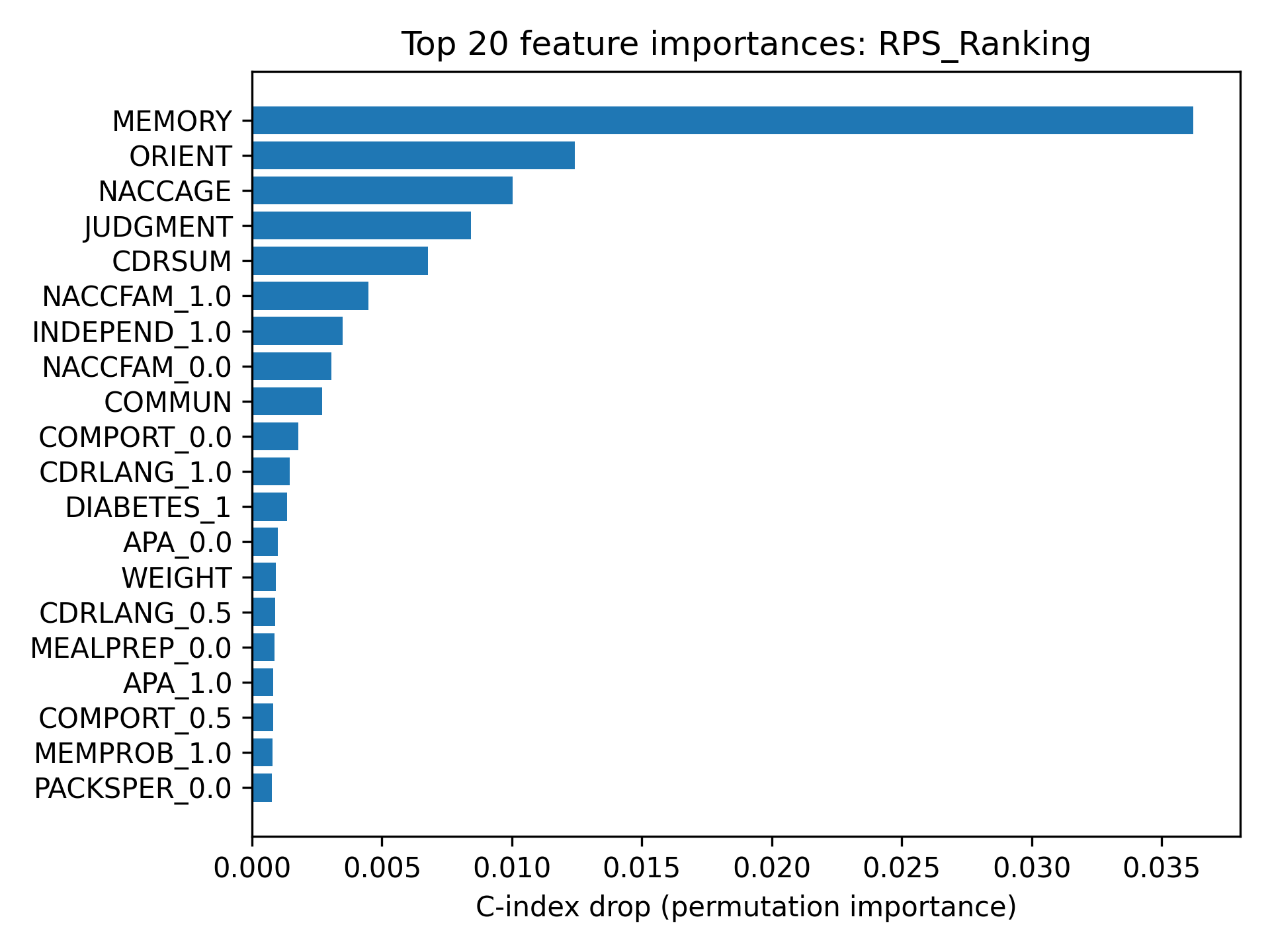}
     \end{subfigure}
     
        \caption{Permutation-based feature importance analysis of DeepHit (top-left), N-MTLR (top-right), RPS (bottom-left), and RPS+Rank (bottom-right).}
        \label{fig:permuation-app}
\end{figure*}

\section{Permutation Importance}
\label{app:perm}

The permutation importance analysis for DeepHit, N-MTLR, RPS, and RPS+Rank can be found in Figure \ref{fig:permuation-app}. We observe that the top five features across all models are identically ordered, including: MEMORY, ORIENT, NACCAGE, JUDGMENT, and CDRSUM. 

Many of the top 20 features that appeared among all models are related to the Clinical Dementia Rating (CDR) exam, which aims to assess the level of impairment a subject faces across a range of tasks. For example, MEMORY, ORIENT, and JUDGMENT quantify impairment in memory, orientation, and problem solving tasks. Additionally, CDRSUM quantifies the total impairment scores over the entire assessment. Other notable features which appear commonly among all experiments include NACCFAM (indicates whether a first-degree family member has cognitive impairment), INDEPEND (level of independence), and MEMPROB (whether the subject personally feels they have more memory problems than most).

\section{KM-Fair Analysis}
\label{app:KM-fair}

\begin{figure*}[ht]
     \centering
     \begin{subfigure}[t]{.8\textwidth}
         \centering
         \includegraphics[width=\textwidth]{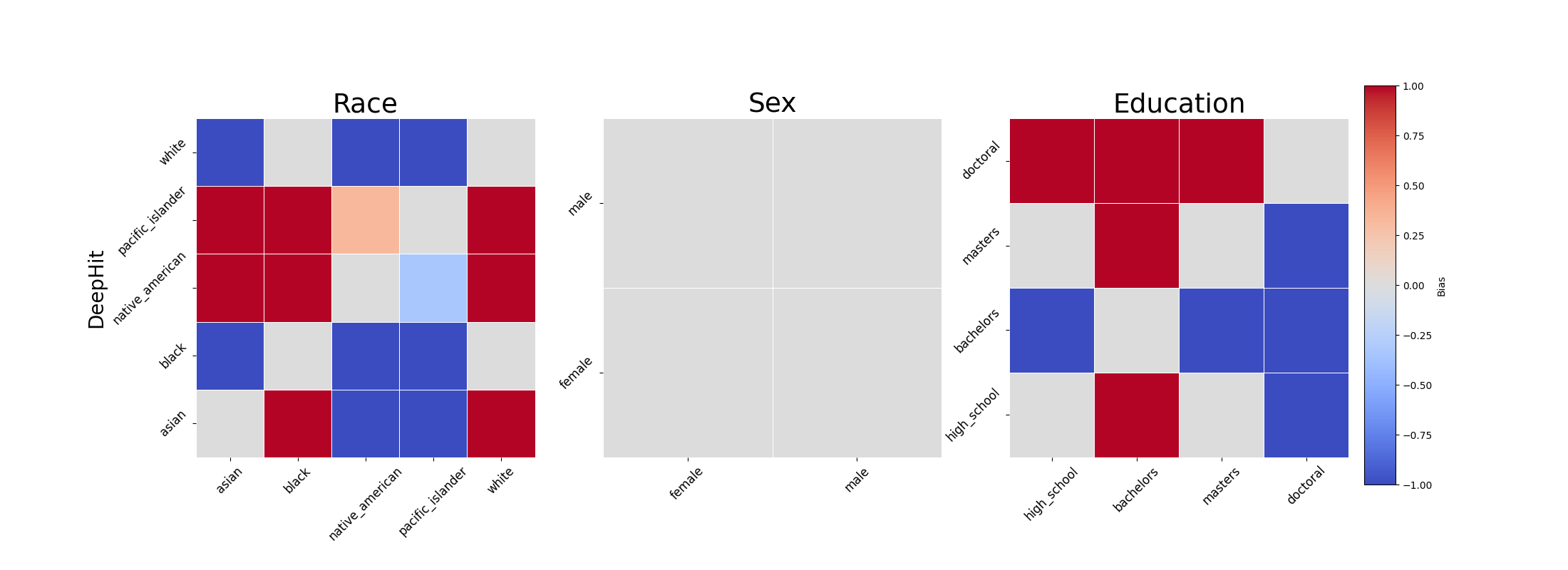}
     \end{subfigure}
     \hfill
     \begin{subfigure}[t]{.8\textwidth}
         \centering
         \includegraphics[width=\textwidth]{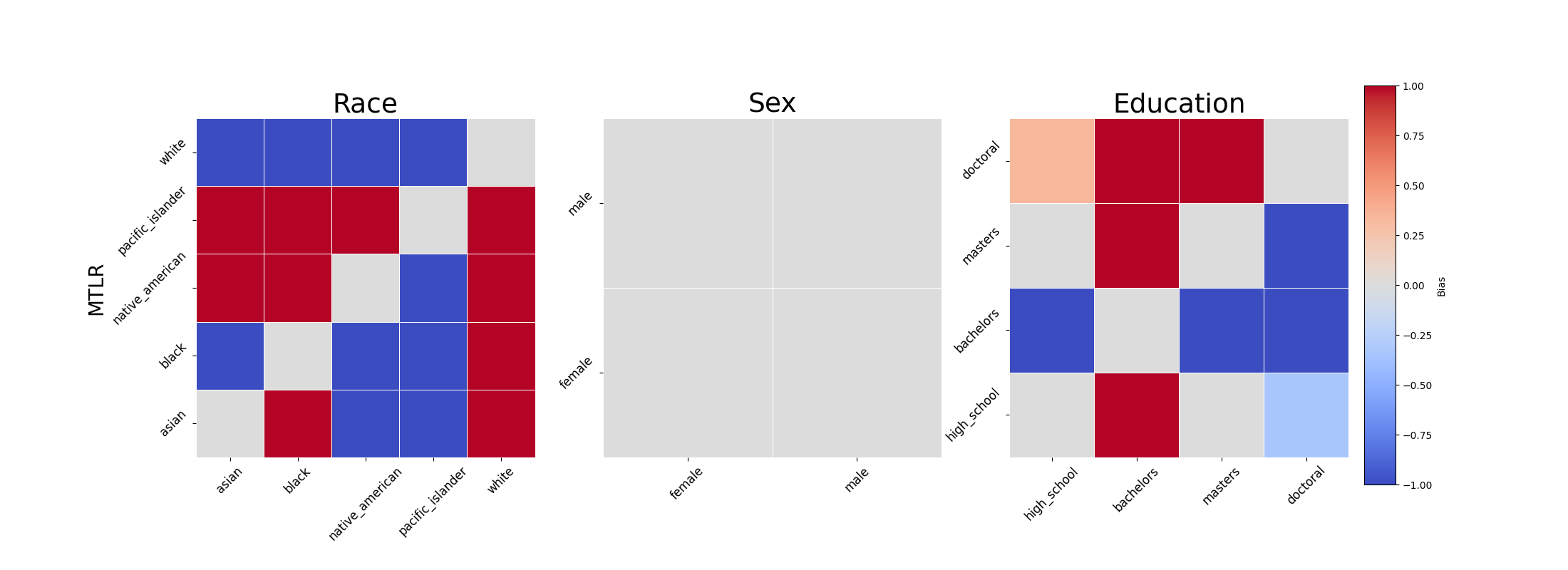}
     \end{subfigure}
     \hfill
     \begin{subfigure}[t]{.8\textwidth}
         \centering
         \includegraphics[width=\textwidth]{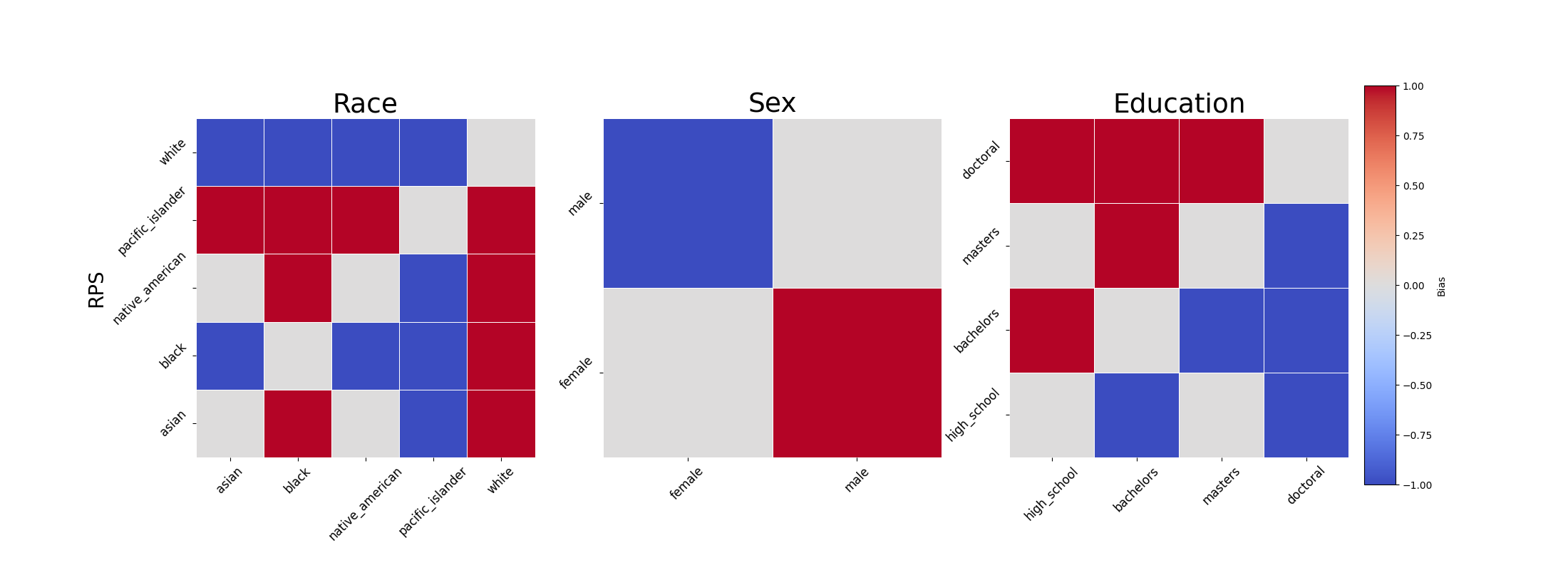}
     \end{subfigure}
     \hfill
     \begin{subfigure}[t]{.8\textwidth}
         \centering
         \includegraphics[width=\textwidth]{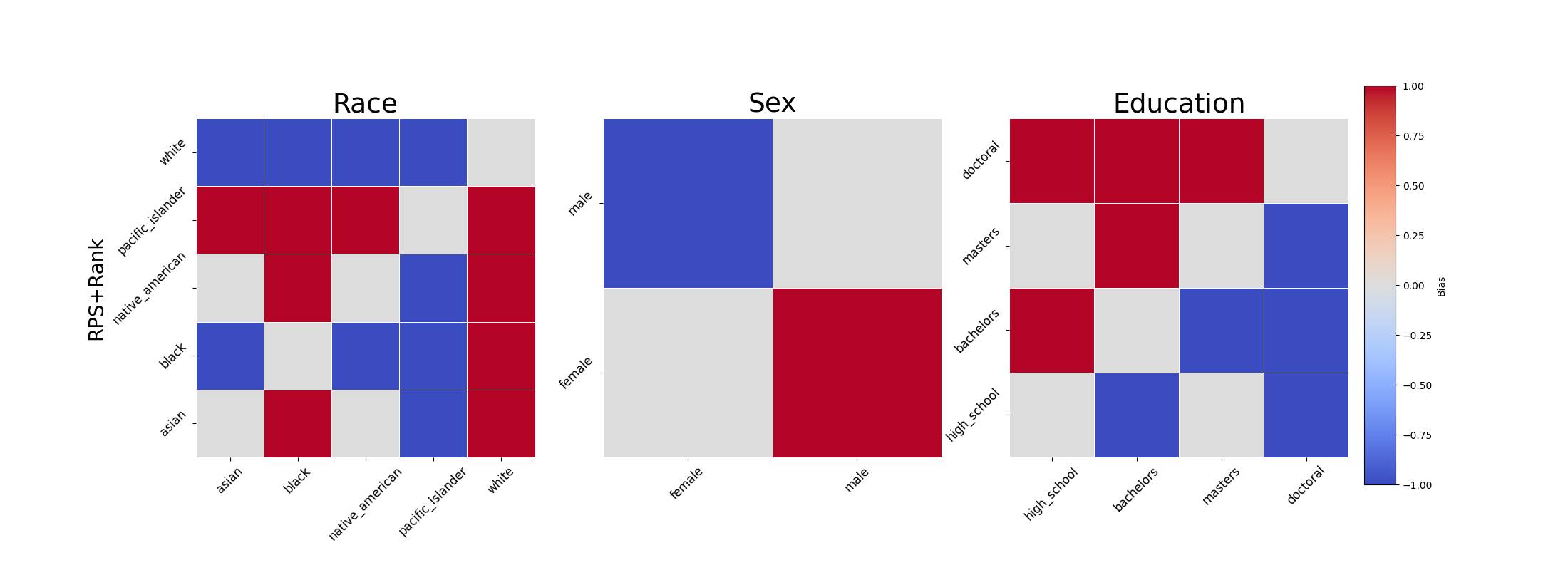}
     \end{subfigure}
     
        \caption{KM-Fair analysis for all models where \colorbox{blue!50}{blue} indicates a model which is biased toward \textit{row attributes} and \colorbox{red}{red} indicates one which is biased toward \textit{column attributes}.}
        \label{fig:km-fair_all}
\end{figure*}

We provide the KM-Fair analysis for DeepHit, N-MTLR, RPS, and RPS+Rank in Figure \ref{fig:km-fair_all}. Here we observe that all models generally exhibit similar bias toward highly represented populations (e.g. white subjects and those without graduate degrees). Interestingly, despite the \textit{sex} category being more or less balanced, NLL, RPS, and RPS+Rank are still biased toward the male population.

% \begin{figure}[t]
%   \centering
%    \includegraphics[width=.9\linewidth]{figures/km-fiar_all.png}

%    \caption{KM-Fair analysis for all models where \colorbox{blue!50}{blue} indicates a model which is biased toward \textit{row attributes} and \colorbox{red}{red} indicates one which is biased toward \textit{column attributes}.}
%    \label{fig:km-fair_all}
% \end{figure}

\end{document}